\pdfoutput=1
\documentclass{article}
\usepackage{iclr2027_conference,times}
\usepackage[T1]{fontenc}
\usepackage{amsmath,amssymb,booktabs,tabularx,array,longtable}
\usepackage{graphicx,microtype,xcolor}
\usepackage{hyperref}
\hypersetup{colorlinks=true,linkcolor=blue,citecolor=blue,urlcolor=blue,
  pdftitle={Recursive Improvement of a Differentiable Scientific Software Ecosystem}}
\title{Recursive Improvement of a Differentiable\\
Scientific Software Ecosystem}
\iclrfinalcopy
\author{Pengcheng Hou$^{1,*}$, Xiaojun Tan$^{2,*}$, Sihan Hu$^{3,1}$,
Ruisi Wang$^{4}$, Shuo Chen$^{5}$, Lei Wang$^{4}$,\\
\bfseries Youjin Deng$^{3,1}$, Kun Chen$^{5,\dagger}$\\[0.6ex]
\normalfont\small $^{1}$Hefei National Laboratory, Hefei 230088, China\\
\normalfont\small $^{2}$Institute for Advanced Algorithms Research, Shanghai 200120, China\\
\normalfont\small $^{3}$University of Science and Technology of China, Hefei, Anhui 230026, China\\
\normalfont\small $^{4}$Institute of Physics, Chinese Academy of Sciences, Beijing 100190, China\\
\normalfont\small $^{5}$Institute of Theoretical Physics, Chinese Academy of Sciences, Beijing 100190, China\\[0.6ex]
\normalfont\small $^{*}$Equal contribution.\quad $^{\dagger}$Corresponding author: \texttt{chenkun@itp.ac.cn}}
\newcolumntype{Y}{>{\raggedright\arraybackslash}X}

\begin{document}
\maketitle
\lhead{Preprint}

\begin{abstract}
    Differentiable programming connects scientific computation with gradient-based inference, learning and design. Extending these capabilities across a heterogeneous software ecosystem requires specialized effort to implement derivatives, integrate interfaces and evaluate quality. AI coding agents can accelerate this transformation, but translating their capabilities into useful scientific software requires identifying research needs and evaluating how well implementations meet them. We present an environment for agent-driven evolution of differentiable scientific software that connects demand identification, development and quality evaluation. A unified differentiation interface exposes reusable derivative rules alongside existing numerical routines, allowing research tasks to share these capabilities. Research requirements guide development, with implementations assessed through independent derivative checks, workflow tests and performance evaluation. Validated software, research programs and tests become shared resources for subsequent studies. We construct and validate automatic differentiation extensions across 20 packages spanning physical, chemical and biological modeling, with research workflows demonstrating reuse across tasks. Benchmarks demonstrate computational savings over finite differences in gradient evaluation and complete parameter estimation. Research-driven revisions make previously unsupported workflows differentiable, correct derivatives of scientific observables and eliminate redundant computation. Quantum-control and thermal-design studies revise objectives in response to physical evaluation, improving designs while reusing existing derivatives. This work provides a practical approach to expanding differentiable programming across established scientific software and organizing AI agents around the recursive improvement of a shared computational ecosystem.
\end{abstract}

\section{Introduction}

Scientific software is a cumulative record of research practice. It preserves
mathematical models, numerical methods and domain conventions as executable
code, allowing new studies to build on established computations. Differentiable
programming extends this reuse from evaluating a model to computing how its
outputs change with parameters, making scientific computations available to
gradient-based learning, inference and design \citep{blondel2024elements}.
The opportunity extends across research fields: reusable derivatives can
connect established numerical implementations to new observations,
objectives and parameterizations without rebuilding a scientific model for
each application.

Realizing this opportunity across a heterogeneous software ecosystem
requires more than applying an automatic differentiation (AD) engine.
Numerical codes expose different
data structures, solvers and execution models, and their derivatives must
preserve domain-specific meaning. A material parameter can affect assembled
equations, the solved field and a measured flux: the derivative must follow
this entire dependency chain. Established AD techniques provide differentiation machinery
\citep{baydin2018,enzyme2020,farrell2013,mitusch2019}; integrating them into
scientific software requires choosing suitable differentiation boundaries,
connecting interfaces and checking numerical results across many useful tools.

AI agents can investigate repositories and carry out
scientific coding tasks \citep{sweagent2024,toolmaker2025,scicode2024,
    scienceagentbench2025,geng2026scienceide}, creating an opportunity to
extend scientific software across domains. Producing code,
however, does not establish which numerical capabilities a research problem
needs or whether an implementation meets those needs. A new observable may
require dependencies that an existing interface omits; even correct
derivatives can optimize a design that fails under physical conditions absent
from its objective. Research programs make these requirements concrete and
provide tests that guide further development.

We ask how agent-driven development and research use can jointly build and
improve a differentiable scientific software ecosystem. Research uses existing
software and returns updated implementations, programs and checks for later
studies, which can reveal further needs. We call this connection
\emph{recursive improvement}.

\begin{figure}[t]
    \centering
    \includegraphics[width=\linewidth]{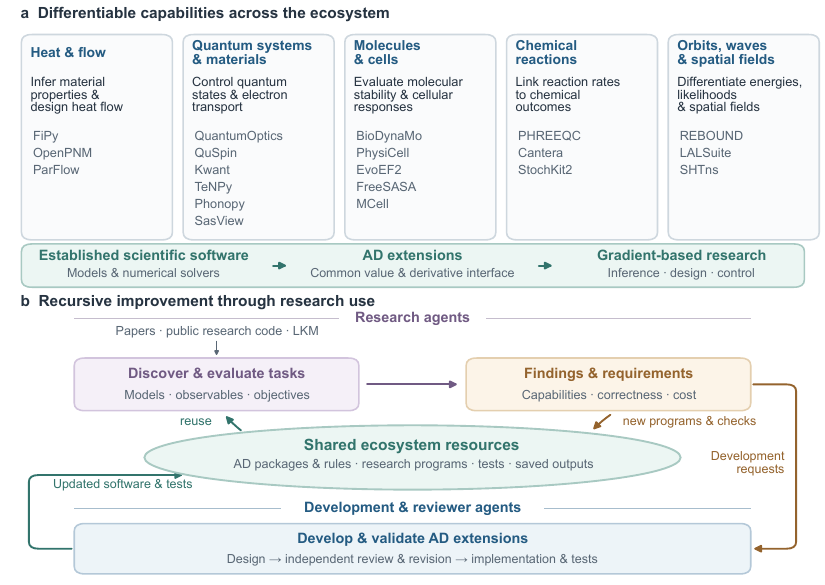}
    \caption{AD extensions and recursive improvement of a scientific
        software ecosystem.
        (a) Extensions connect established computations to gradient-based
        research across 20 packages, using runtime-specific
        value-and-derivative interfaces.
        (b) Research reuses shared resources and returns programs, checks
        and development requirements. Updated software and tests then
        support further research. Papers, public research code and LKM inform
        task discovery (Section~\ref{sec:research-feedback}).}
    \label{fig:ecosystem}
\end{figure}

We construct an ecosystem comprising scientific software and its AD
extensions, common differentiation conventions, executable research programs
and numerical references (Figure~\ref{fig:ecosystem}). The evaluated collection
spans molecular and cellular modeling, chemical kinetics, materials and
quantum physics, and Earth and space sciences. Research programs compose the
extensions' values and derivative actions with their own inputs and objectives,
allowing the same numerical capability to serve different studies.

Our contributions are:
\textbf{(1) A differentiable scientific software ecosystem.}
We construct and validate AD extensions across 20 scientific packages, with
common differentiation conventions, reusable research programs and numerical
references. Evaluations demonstrate reuse across tasks and computational
utility (Section~\ref{sec:evaluation}).
\textbf{(2) Research-driven recursive improvement.}
We show how research feedback expands derivative capabilities, corrects
derivatives of scientific observables and removes redundant computation.
Research inputs, results and checks are reused to validate successive software
revisions
(Sections~\ref{sec:ecosystem} and~\ref{sec:recursive-results}).
\textbf{(3) Evolving research applications.}
We demonstrate how reusable derivatives support research applications that
develop in response to physical evaluation, including more error-tolerant
quantum control and thermal designs for additional heating conditions.
Prepared quantum states also support a subsequent sensing study
(Section~\ref{sec:evolution}).

\section{A differentiable scientific software ecosystem}
\label{sec:ecosystem}

The environment connects reusable numerical capabilities to research programs
through development, evaluation and revision (Figure~\ref{fig:ecosystem}).

\subsection{Reusable computations and differentiation interfaces}

An AD extension exposes values and derivatives for specified continuous
inputs within a documented domain. It uses original APIs and numerical
routines, or kernels checked against original outputs. For $y=f(\theta;c)$,
$\theta$ denotes active inputs and $c$ fixed context such as topology or a
numerical branch.
Jacobian--vector products (JVPs) and vector--Jacobian products (VJPs) describe
the numerical response; research code defines the task's parameterization,
observations and loss.
Different objectives can therefore share a derivative implementation
\citep{blondel2022}. Our diffusion rules for FiPy \citep{guyer2009}
differentiate the conductivity-to-temperature solve: inversion compares the
field with observations, whereas cloak design measures its exterior disturbance.

The shared abstraction has runtime-specific bindings, not a single execution
engine. Our Python ChainRules
package registers explicit rules for existing callables, following Julia
ChainRules' separation of rules from AD engines \citep{julia_chainrules}.
The evaluated FiPy programs use custom Autograd rules \citep{autograd};
QuantumOptics \citep{kramer2018quantumoptics} uses ChainRulesCore with Zygote
\citep{zygote2018}. Appendix~\ref{app:protocol-bindings} gives the bindings
and a concrete task-to-rule call. Research programs, numerical checks and
saved outputs complement these packages as resources for later studies.

\subsection{Development guided by research use}
\label{sec:research-feedback}

Development agents inspect the original software
and propose computations to extend, specifying the APIs, active inputs and
differentiation strategy.
Separate reviewer agents assess the design; developers revise it and then
implement the rules, examples and numerical tests
(Appendix~\ref{app:construction}).

Research agents turn scientific questions from papers, public research code
and original-software examples into models, objectives and executable
evaluation programs. Alongside these sources, the Large Knowledge Model (LKM)
\citep{bohrium_lkm} represents scientific literature as source-grounded
reasoning graphs linking research questions, workflows, conclusions and
supporting evidence. Evaluation checks original-value agreement, derivatives against
independent references, and scientific outcomes under specified conditions;
computational costs are measured separately.

Research feedback carries a computation, fixed inputs and numerical checks into
a development request. In Kwant, the request supplied an eight-gate device and
required gate-to-transmission derivatives checked against original values,
finite differences and flux conservation. Development returned a package rule;
research-side retests used the same device, then reused its optimized design
to test a second implementation update (Section~\ref{sec:transport-design}).
Physical evaluation can instead prompt revised research objectives using existing
derivatives. Software, programs, checks and saved results remain available to
subsequent tasks.

\section{Capabilities and scientific utility}
\label{sec:evaluation}

We evaluate available capabilities, numerical validity and computational utility
across the 20-package ecosystem. Table~\ref{tab:capability-reuse} links
representative computations to their research uses and checks.

\subsection{Reusable capabilities and numerical validation}

Shared solves, propagations and rule compositions support different task
definitions, from quantum control with QuSpin \citep{weinberg2017,weinberg2019}
to porous transport with OpenPNM \citep{gostick2016openpnm} and hydrologic
sensitivities with ParFlow \citep{kuffour2020parflow}. Task code selects the
observable or objective; the package supplies the reusable derivative.

\begin{table}[!htbp]
\centering
\small
\caption{Research uses and checks of reusable derivative capabilities.
Rows distinguish optimization studies, composed-workflow checks and sensitivity
cases. Workflow counts are reported evaluations; definitions and sources
appear in Appendices~\ref{app:sources}--\ref{app:software-revisions}.}
\label{tab:capability-reuse}
\setlength{\tabcolsep}{4pt}
\renewcommand{\arraystretch}{1.12}
\begin{tabularx}{\linewidth}{@{}>{\raggedright\arraybackslash}p{0.22\linewidth}
>{\raggedright\arraybackslash}p{0.36\linewidth}Y@{}}
\toprule
Numerical capability & Research computations using it & Validation evidence\\
\midrule
Diffusion sensitivities\newline FiPy
& Temperature-based conductivity inversion; thermal-cloak optimization under different heating objectives.
& Native field replay and finite-difference gradient checks in both studies (Appendices~\ref{app:cloak},~\ref{app:inverse}).\\
\addlinespace[3pt]
State-propagation derivatives\newline QuSpin
& Nominal and coupling-aware state-preparation optimizations: one propagation rule, 96 controls.
& Independent matrix-exponential replay and directional finite differences (Appendix~\ref{app:quantum-control}).\\
\addlinespace[3pt]
Evolution, steady-state and spectral rules\newline QuantumOptics
& Value and gradient checks of composed control, inference and spectral objectives.
& 14 composed cases pass reference-value and finite-difference checks (Appendix~\ref{app:workflow-expansion}).\\
\addlinespace[3pt]
Transport and flux sensitivities\newline OpenPNM
& Pressure-driven flow, diffusion, heat, reaction--diffusion and advection--diffusion sensitivities.
& 66/80 pass value/derivative checks; 14 hard-threshold invasion cases skipped (Appendix~\ref{app:workflow-expansion}).\\
\addlinespace[3pt]
Simulator-connected sensitivities\newline ParFlow
& Pressure and saturation sensitivities; groundwater-head and water-storage objective gradients.
& 80 sensitivity cases pass original-simulator finite-difference checks (Appendix~\ref{app:broader-capabilities}).\\
\bottomrule
\end{tabularx}
\end{table}

EvoEF2 \citep{huang2020evoef2} prepares a molecular structure before
differentiating continuous scoring weights, leaving discrete preparation choices
outside the derivative. Appendix~\ref{app:broader-capabilities} gives the
wider task families and numerical domains.

Checks use original-software or independent values, finite-difference (FD)
or analytic derivatives, and independent task-level comparisons. For the
QuantumOptics 144-control steady-state calculation, the full gradient has
relative $L_2$ discrepancy $1.16\times10^{-8}$ against central FD. A
FiPy scaling series reaches 1,024 conductivity parameters
with maximum relative gradient discrepancy $8.46\times10^{-8}$
(Appendix~\ref{app:gradient-scaling}). Task-specific definitions and tolerances
are given in Appendix~\ref{app:sources}.

\subsection{Derivative costs and research-task efficiency}

\begin{figure}[!htb]
    \centering
    \includegraphics[width=\linewidth]{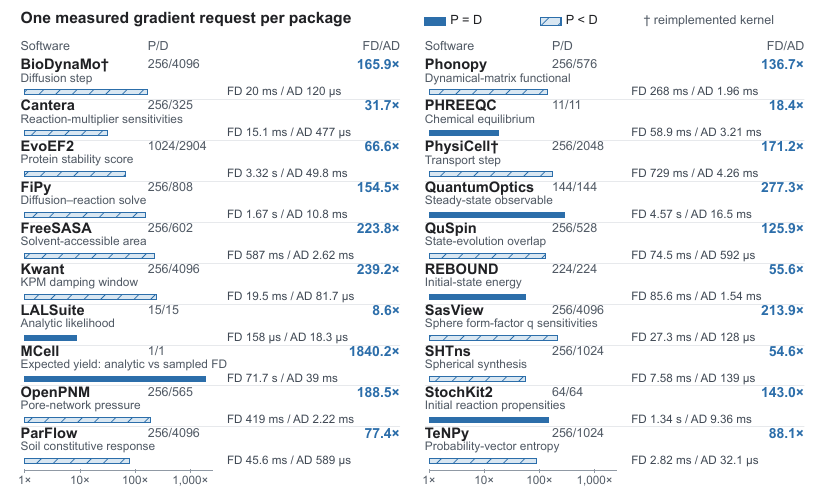}
    \caption{Computational utility of derivative implementations across 20
    scientific packages. Each row measures a specified gradient workload.
    The component column gives $P/D$: FD evaluates $P$ derivatives,
    while AD returns a gradient over all $D$ independent parameters. Solid
    bars have $P=D$; hatched bars have $P<D$. Bars show ratios of median
    runtimes, $t_{\rm FD}(P)/t_{\rm AD}(D)$, with absolute times below.
    $\dagger$: reimplemented diffusion kernels checked against original-software
    outputs. MCell compares an analytic expectation rule with native Monte
    Carlo FD under a 10\% gradient-accuracy criterion. Workload definitions
    and repeats appear in Table~\ref{tab:portfolio-workloads}; validation
    details appear in Appendices~\ref{app:additional-profiling}
    and~\ref{app:mcell-rate-amplitude}.}
    \label{fig:portfolio}
\end{figure}

Figure~\ref{fig:portfolio} compares AD and central FD costs for specified gradient
requests, from local observables to numerical solves. AD is less costly in all 20
measured requests. The full
144-control QuantumOptics gradient, for example, takes 16.5\,ms with AD and
4,570\,ms with FD. AD can share work across gradient components, whereas
central FD repeats forward evaluations at perturbed parameters
\citep{baydin2018}. The plotted Kwant \citep{groth2014} request differentiates a spectral-window
operation; the scattering-based filter studied in Section~\ref{sec:transport-design}
is an additional capability with a separate cost measurement.

Fourteen rows measure FD on fewer than all active coordinates; the plotted
times are not extrapolated to a full gradient. For MCell \citep{husar2024mcell},
a model-specific expected-yield rule avoids the native Monte Carlo sampling
required by FD at the prescribed accuracy
(Appendix~\ref{app:mcell-rate-amplitude}).

\label{sec:inverse-problems}
We next ask whether a cheaper gradient improves a complete research task.
In matched synthetic inversions, FiPy recovers 32 regional conductivities
from temperature fields under two thermal loadings, and Kwant recovers 32
onsite potentials from local densities of states (LDOS) at four energies. AD and FD
use the same noiseless, same-model observations and bound-constrained
L-BFGS-B optimizer \citep{zhu1997lbfgsb}. Both recover parameters with
similar accuracy and make similar predictions under an additional FiPy
loading or at additional Kwant energies. In representative fits,
held-out temperature root-mean-square error is $1.72\times10^{-7}$ for AD
and $1.34\times10^{-7}$ for FD; the corresponding withheld LDOS errors
are $3.87\times10^{-7}$ and $3.89\times10^{-7}$. Median optimization time
is lower with AD by factors of 39.5 and 29.5, respectively
(Appendix~\ref{app:inverse}).

\section{Research-driven evolution of shared capabilities}
\label{sec:recursive-results}

Research use reveals requirements beyond the initial derivative implementation.
We trace how task definitions and computational demands lead to new capabilities,
numerical repairs and execution improvements, with research inputs and outputs
reused to test revisions.

\subsection{Iterative software improvement through research tasks}
\label{sec:transport-design}

An electronic energy filter should suppress transmission in one energy
window while retaining it in another \citep{zhouyin2023}. We tune eight
gates in a $24\times6$ Kwant device connected to two leads. The initial
extension differentiated supplied scattering matrices but could not carry
gate changes through the solve that produces transmission. The research
task specified the missing gate-to-transmission path, together with value,
derivative and flux-conservation checks. A new implicit scattering rule
passed an independent task-level retest and made the filter objective
differentiable.

\begin{figure}[!htbp]
    \centering
    \includegraphics[width=\linewidth]{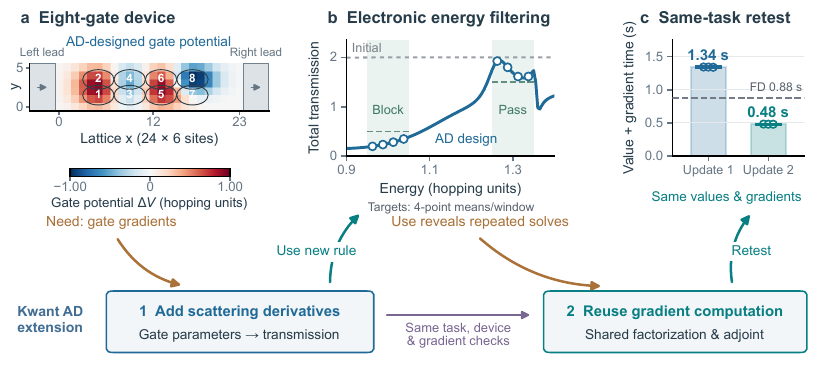}
    \caption{A research task drives a new derivative capability and then
        an implementation improvement in Kwant.
        (a) Eight-gate potential designed using the first scattering-rule update.
        (b) Native transmission of initial and AD-designed devices at 101
        energies; circles mark four reserved points per energy window.
        Dashed thresholds apply to their mean transmission within each
        window. Two incoming channels give a maximum of two.
        (c) Median time and all three process measurements for one value and
        full-gradient request at the same saved design, with matching
        numerical results; the gray line is the median FD reference for the
        same request. The feedback band connects the missing derivative,
        its use in design and the later execution improvement.}
    \label{fig:recursive-evidence}
\end{figure}

Using that rule, an AD-based design meets the predeclared thresholds at
eight energies reserved from the six used for optimization: mean stop/pass
transmissions are 0.266/1.737, against limits of at most 0.5 and at least
1.5. FD finds a similar design (Appendix~\ref{app:transport-design}).
Figure~\ref{fig:recursive-evidence}a--b shows the saved potential and native transmission.

Repeated gradient evaluations revealed that the first implementation rebuilt
and factorized the same scattering system for each gate direction. The next
revision shared the linearization and added an adjoint. At both the
original and saved-design inputs, values and full gradients remain matched;
for one six-energy value-and-gradient request, derivative factorizations
fall from 48 to six. At the saved design, median time falls from 1.34 to
0.48\,s, compared with 0.88\,s for central FD, in three processes per method
(Figure~\ref{fig:recursive-evidence}c). These timings compare fixed inputs,
not complete design runs. The saved device and numerical checks are reused
to validate the revised gate-to-transmission rule: the design result also
supports a subsequent software improvement
(Appendix~\ref{app:transport-design}).

\subsection{Research-driven improvements across domains}
\label{sec:reuse}

StochKit2 \citep{sanft2011} shows why internally consistent derivatives can
still describe the wrong scientific quantity. A research program measures
how reaction rates change a terminal-population histogram with specified
bins. In a two-sample regression test, the requested probability vector is
$[1,0]$ and its reference derivative is $[0,0]$, yet the old rule returns
$[-0.5,0.5]$. Both derivative modes agree because both recompute automatic
bin edges instead of using those requested by the research program.

The repair carries the requested bins through both modes and agrees with an
independent reconstruction from native trajectories. The bin-specific test
then remains available during later changes, including a revision that
reduces duplicate native calls from four to two with bitwise-identical
outputs. In the native 512-trajectory screen, one ensemble still misses
a low-count-bin analytic-gradient criterion: preserving the requested
observable and achieving sufficient sampling precision are separate
requirements (Appendix~\ref{app:stochastic-followup}).

Workflow reports describe further capability changes. An OpenPNM patch adds
reaction-source and directed-flow derivatives, bringing
reaction--diffusion and advection--diffusion into its checked transport paths.
BioDynaMo repairs mechanics updates to preserve the original simulation's
meaning. QuSpin brings propagation derivatives
from research programs into a shared package rule, which later supplies both
control objectives in Section~\ref{sec:quantum-control}. The workflow reports
and subsequent uses in Appendix~\ref{app:workflow-expansion} distinguish
capability expansion, numerical repair and consolidation into reusable software.

Across these cases, research definitions become numerical checks, and saved
inputs and outputs support later software changes and applications.

\section{Evolving research applications}
\label{sec:evolution}
\label{sec:applications}

Research applications develop by refining tasks and reusing their results.
Here existing quantum-propagation and diffusion derivatives support design
revisions guided by physical evaluation. Correct gradients alone do not
ensure that a design works under the conditions of interest. Prepared
quantum states also become inputs to a subsequent sensing study.

\subsection{Robust quantum control and phase sensing}
\label{sec:quantum-control}

\begin{figure}[!t]
    \centering
    \includegraphics[width=\linewidth]{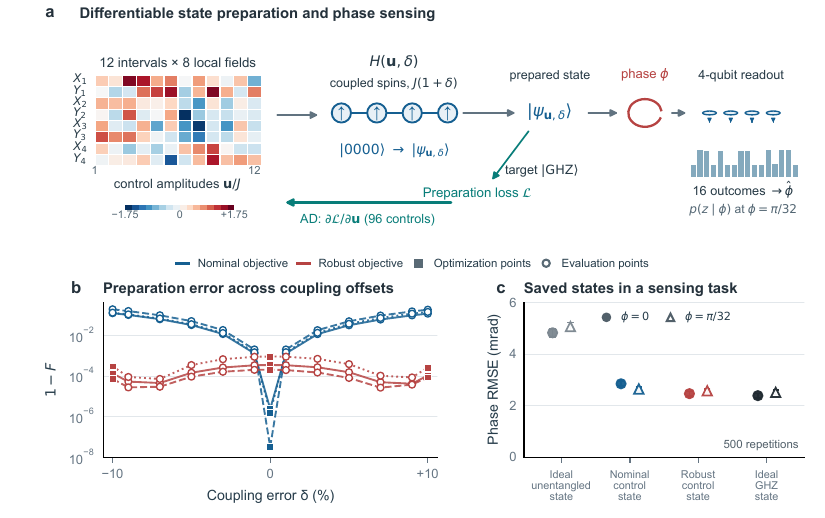}
    \caption{A revised control objective improves robustness using the same
        QuSpin derivatives; saved states then support a measurement study.
        (a) Gradients pass through evolution to 96 controls. The pulse map
        shows the selected robust control. Readout probabilities use
        coupling shift $\delta=0.1$ and encoded phase $\phi=\pi/32$.
        (b) Error $1-F$ for three paired initializations, where $F$ is GHZ
        squared overlap. Line styles identify pairs; squares mark optimization
        conditions and circles other evaluations.
        (c) A subsequent sensing task reuses two selected, unpaired states.
        Phase-estimation root-mean-square error (RMSE) is evaluated at known
        $\delta=0.1$ and ideal reference probes. Each point uses 500
        repetitions of 10,000 measurements. Markers distinguish phases;
        bars show approximate Monte Carlo standard errors. Reference probes
        have no matched preparation cost. Protocols appear in
        Appendices~\ref{app:quantum-control} and~\ref{app:quantum-metrology}.}
    \label{fig:evolution}
\end{figure}

Entanglement can improve phase sensitivity \citep{giovannetti2011metrology},
provided suitable probes can be prepared with the available controls
\citep{yang2020probe}. We use QuSpin
propagation and its derivatives to prepare a four-spin
Greenberger--Horne--Zeilinger (GHZ) state, an equal superposition of all-up
and all-down spins. Local fields along two transverse axes act on each spin
over 12 intervals, giving 96 bounded amplitudes. The same propagation
derivative supplies the gradient for nominal and coupling-aware objectives
(Figure~\ref{fig:evolution}a); only the research objective and its evaluation
conditions change. Differentiable quantum control and ensemble objectives
provide the relevant methodological context
\citep{leung2017,schaefer2020control,goerz2022semiad,chen2014ensemble}.

Fifteen initial nominal designs have fidelity above 0.99999 at the intended
coupling $J$, yet replaying fixed pulses with a common coupling shift of up
to $\pm10\%$ exposes sensitivity. Here $\delta$ denotes the fractional
change in $J$. The revised objective averages preparation
error at three coupling values. At $T=4/J$ with $\hbar=1$, three paired
initializations compare the objectives with equal budgets of 660
single-scenario value-and-gradient evaluations. On ten additional offsets
between the training values, the worst error $1-F$ falls from
10.2--15.8\% to 0.020--0.090\% (Figure~\ref{fig:evolution}b). This gain
comes with larger root-mean-square control amplitudes and a small loss of
nominal fidelity; all six runs stop at their prescribed budgets. The
revised objective therefore yields controls that better tolerate coupling
errors in this four-spin task, using the existing propagation derivative.

We next reuse saved states as probes to estimate an unknown collective spin
rotation from four-bit measurement outcomes, without another pulse optimization. We
select one state from each control objective at a fixed, known $+10\%$
coupling error and use the same ideal readout and estimator. Root-mean-square
errors are 2.48--2.86 milliradians across two phases. Ideal unentangled and
GHZ probes yield 4.84--5.08 and 2.40--2.53 milliradians, respectively
(Figure~\ref{fig:evolution}c). Both saved states retain sensing utility
close to the ideal GHZ reference under this readout. This subsequent-use test
is separate from the paired control comparison
(Appendix~\ref{app:quantum-metrology}).

\subsection{Thermal-cloak design under multiple heating conditions}

\begin{figure}[!htbp]
    \centering
    \includegraphics[width=\linewidth]{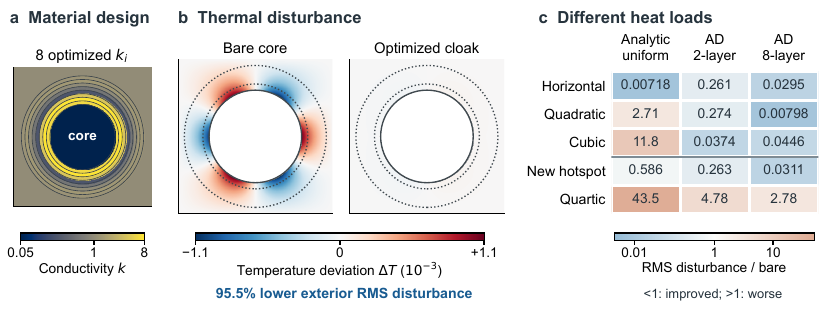}
    \caption{A thermal cloak designed with FiPy diffusion derivatives
        following a joint objective and training-grid revision.
        The eight-layer design is optimized on a $192\times192$ grid and
        evaluated at $576\times576$ with material parameters fixed.
        (a) AD-optimized conductivities. (b) Cubic-load temperature deviations
        from the homogeneous reference, on a shared scale. Interiors are masked; dotted
        circles mark the observation annulus.
        (c) RMS disturbance relative to the bare core. The line separates
        three design loads from two new loads. Ratios below one (blue) indicate
        improvement; above one (orange), greater disturbance. Conductivity
        and ratio color scales are logarithmic. Both optimized designs use
        AD; the uniform coating is an analytic reference.}
    \label{fig:research-outcomes}
\end{figure}

A thermal cloak redirects heat so that an inclusion leaves little signature
in the surrounding temperature field \citep{han2013cloak}. A design that
works under one heating pattern can remain visible under another.
We use FiPy's
implicit diffusion derivatives to choose conductivities for eight concentric
layers around a low-conductivity core, under material bounds and a budget.
The gradient connects the layer conductivities to exterior temperature
disturbance after the steady diffusion solve
(Figure~\ref{fig:research-outcomes}a--b). It is the same class of diffusion
capability used for conductivity inversion in
Section~\ref{sec:inverse-problems}, here composed with a design objective.

An initial design trained on two boundary-heating patterns increases, rather
than suppresses, the disturbance under a cubic pattern. The revised task
adds that load to its objective and trains on a finer grid, retaining the
diffusion rule. After this joint revision, on a common $192^2$ evaluation
grid, cubic root-mean-square
(RMS) disturbance relative to the bare core falls from 2.96 to 0.00371;
the revised ratio is 0.0446 at $576^2$ (Appendix~\ref{app:cloak}). Figure~\ref{fig:research-outcomes} shows the
revised design's performance under its design loads and additional heating
conditions.

For a fixed revised design evaluated at $576^2$, disturbance falls by
95.5--99.2\% on the three design loads and by 96.9\% under a new localized
hotspot. A new quartic profile instead gives 2.78 times the bare-core
disturbance (Figure~\ref{fig:research-outcomes}c); its grid-refinement response
fails the prespecified stability check (Appendix~\ref{app:cloak}). The
analytic uniform coating and an optimized two-layer design each achieve
lower disturbance on individual loads. The revised design is a
multiload compromise. The physical tests identify both where that design is
effective and which conditions remain outside its demonstrated performance.

\section{Related work}

\paragraph{Differentiable scientific computing.}
Established AD systems provide operator overloading, source transformation
and compiler-based differentiation \citep{griewank1996,hascoet2013,enzyme2020}.
Composable transformations and rules at numerical-solver boundaries support
differentiable simulation and scientific modeling
\citep{jax2018,difftaichi2020,schoenholz2020,farrell2013,mitusch2019,
    blondel2022,rackauckas2020universal}. ChainRules separates derivative
definitions from AD engines; GradBench and earlier benchmarks compare
implementations across languages and workloads
\citep{julia_chainrules,gradbench,srajer2018}. We build on these techniques
to extend established scientific packages. We examine how the resulting
derivatives preserve scientific meaning across research uses, and how
task-level requirements change shared implementations and their checks.

\paragraph{Agents for software and scientific research.}
Agent systems and benchmarks study repository repair
\citep{swebench2024,sweagent2024,openhands2025}, scientific coding and
research extensions
\citep{scicode2024,ainsteinbench,swescience,scienceagentbench2025,rexbench2026},
and machine-learning engineering \citep{mlebench2025}.
ToolMaker builds agent-callable tools from papers with code
\citep{toolmaker2025}; Lahlou et al.\ translate a legacy Fortran model into
JAX with a numerical reference \citep{lahlou2026}. ScienceIDE uses scientific
repositories as environments for agent learning and evaluation
\citep{geng2026scienceide}. Our studies follow reusable numerical software
across research tasks: requirements revealed in use change its implementation,
while validated derivatives and saved outputs support subsequent studies.

\paragraph{Evaluator-driven and recursive improvement.}
FunSearch and AlphaEvolve use evaluators to guide program search
\citep{funsearch2023,alphaevolve2025}; the Darwin G\"odel Machine modifies
the agent itself \citep{zhang2026darwin}. Our object of improvement is
scientific software together with the research programs that use it.
Evaluation can change the numerical capability or the scientific objective,
and a preceding research result can become an input to later software
validation. The cases examine these connections within a shared ecosystem.

\section{Conclusion}

We have built a differentiable scientific software ecosystem through
agent-driven construction and research-guided improvement. Across 20
scientific packages, numerical checks
validate diverse derivative computations and measured gradient requests
cost less than their finite-difference comparisons. The research studies
show how these capabilities support inference and design, and how feedback
adds missing dependencies, corrects scientific semantics and reduces
repeated computation in shared implementations.

The ecosystem develops its numerical capabilities and their uses together.
Scientific checks determine whether a computation matches the intended
observable; physical evaluation tests whether that computation serves the
research objective under the conditions of interest. These different
questions motivate changes to shared implementations or to the objectives
built on them. Executable research programs, checks and saved outputs then
provide inputs for subsequent research and software validation. Recursive
improvement connects individual task outcomes to the continuing development
of these shared resources.

The broader opportunity is to use AI agents to reshape how scientific
software ecosystems develop and combine their capabilities. Differentiation
provides a common operation through which many established computations
can participate in learning, inference and design. Similar connections
between development and research could extend other capabilities, including
uncertainty quantification and coupled simulation. Accumulating implementations
and the scientific knowledge needed to validate them offers a route from
individual software advances to resources that support a wider range of
research.

\section*{Reproducibility statement}
Appendix~\ref{app:construction} describes the agent development procedure and
differentiation interfaces. Appendix~\ref{app:sources} specifies the evaluated
computations, numerical checks and timing procedures.
Appendix~\ref{app:software-revisions} details software revisions and retests;
Appendices~\ref{app:quantum-control}--\ref{app:inverse} give research-task
models and settings. Numerical records, componentwise AD/FD comparisons,
task programs and portable verification checks are provided in the
supplementary archive. It records the relevant software versions and runtime
requirements, and distinguishes reanalysis of saved records from fresh
scientific reruns with package-specific environments. The paper and appendices
define the reported inputs, comparisons and evaluation conditions.

\section*{AI use statement}
The development-agent setup is described in Appendix~\ref{app:construction}.
Generative AI tools supported software investigation, AD-interface design,
implementation, testing and revision. They also assisted literature discovery,
study and experiment design, execution and analysis, and manuscript drafting,
revision and figure preparation. Synthetic observations were generated by
numerical simulation programs developed with agent assistance. Validation
procedures included package tests and comparisons with independent numerical
references. The authors take responsibility for the study and the final text,
methods, results and claims.

\section*{Acknowledgments}

Y.D. and K.C. are supported by the Strategic Priority Research Program of the Chinese Academy of Sciences under Grant No. XDB1680102.
P.H., S.H., and Y.D. are supported by the National Natural Science Foundation of China (under Grant No. 12275263) and the Quantum Science and Technology-National Science and Technology Major Project (under Grant No. 2021ZD0301900).
R.W. and L.W. are supported by the National Natural Science Foundation of China under Grants No. T2225018, No. 92270107, and No. 12188101, No. T2121001, and the Strategic Priority Research Program of Chinese Academy of Sciences under Grants No. XDB0500000 and No. XDB30000000.
S.C. and K.C. are supported by the National Key Research and Development Program of China under Grant
No. 2024YFA1408604 and the National Natural Science Foundation of China under Grants No. 12474245
and No. 12447103.

\bibliography{references}
\bibliographystyle{iclr2027_conference}

\clearpage
\appendix
\section{Construction, interfaces and research evaluation}
\label{app:construction}

\subsection{Development roles and stage outputs}

The development procedure used for the AD extensions separates investigation and specification,
independent design review, revision, and implementation
(Table~\ref{tab:construction-stages}). The reviewer reconstructs the original
interface inventory and checks a working runtime in a separate checkout;
the developer revises the specification before implementation. Agents
generate the software-specific configurations, derivative designs, code and
tests. The development agents used Codex with the \texttt{gpt5.6sol} model.
The research team selects software, establishes the common procedure,
initiates runs and provides computing resources.

Saved prompts, target configurations and preceding outputs connect the stages.
The runner also offers a separately invoked delivery review. This optional
stage and repeated design reviews are not documented for every package.
The code and data guide identifies the runner and prompt versions.

In one retained QuSpin construction, the specification stage produced a
proposed design, a separate review requested changes, and the revision stage
passed an explicit target list to implementation. For
\texttt{hamiltonian.evolve}, a review finding required initial-state
derivatives to be distinguished from derivatives of the Hamiltonian itself.
The revised target fixed the Hamiltonian and time grid, differentiated the
initial state, and required original-value agreement, multistep directional
checks and JVP/VJP duality. The later Hamiltonian-active propagation used
in Section~\ref{sec:quantum-control} is a distinct extension. This example
shows how a review changes a numerical design, rather than only approving
an API name.

\begin{table}[htbp]
    \centering
    \small
    \caption{Inputs and outputs of the four-stage development procedure.
        The reviewer receives the proposed design but inspects the original software
        independently. Each stage passes its documented outputs to the next.}
    \label{tab:construction-stages}
    \begin{tabularx}{\linewidth}{@{}lYY@{}}
        \toprule
        Stage          & Inputs                                                                      & Output used by the next stage \\
        \midrule
        Investigation  & Original software, generic development instructions, working runtime
                       & Software-specific specification, numerical targets and runtime observations                                 \\
        Design review  & Proposed specification and separate original-software checkout
                       & Findings on scope, numerical semantics and differentiation route                                            \\
        Revision       & Specification and review findings
                       & Revised design and implementation targets                                                                   \\
        Implementation & Revised design, original code and numerical references
                       & AD extension, documentation, package tests and explicit unresolved cases                                    \\
        \bottomrule
    \end{tabularx}
\end{table}

In the retained QuSpin construction, implementation reports 128 of 130
targets; two reproducible original-call failures remain unresolved.
Source snapshots bind the saved prompts, targets and outputs. The model is
author-confirmed; the available run metadata do not retain per-run reasoning
settings or total development budgets. The records support the documented
handoffs and design decisions, rather than a controlled comparison of agent
development efficiency.

\subsection{Workflow discovery and evaluation}
\label{app:discovery}

Research agents use scientific literature, public research code and
original-software examples to construct executable tasks and evaluate their
numerical results and costs. Alongside these sources, the Large Knowledge
Model (LKM) provides source-grounded reasoning graphs linking research
questions, workflows, conclusions and supporting evidence. The original code
and specified inputs, rather than the reasoning graph alone, define the
computation.

The detailed cases illustrate complementary forms of rule reuse, composition,
numerical repair and execution changes. They were selected retrospectively,
so their counts do not estimate a development success rate. Within the control, material-design
and transport studies, objectives, budgets and evaluation conditions are fixed
before the corresponding execution. Regression tests check repaired behavior;
conditions not used to guide a revision assess subsequent use.

Feedback specifies the affected computation, the missing derivative or observed
defect, and the checks required of a revision. The revised package is exercised
through the research program; implementation-cost comparisons reuse fixed
inputs and compare numerical outputs before timings. The two Kwant updates in
Appendix~\ref{app:transport-design} illustrate this sequence, with retained
requests, source changes and independent retests.

\subsection{Differentiation interfaces used in the experiments}
\label{app:protocol-bindings}

The Python ChainRules implementation associates explicit derivative rules
with callables. A rule accepts named input tangents or active parameters and
returns a value with a tangent or a pullback; the package does not trace
arbitrary Python code. It is separate from Julia's ChainRulesCore.
The QuSpin construction and control experiments use the recorded Python
ChainRules 0.1.0 binding; the Kwant inverse-problem archive uses 0.1.1.dev0.

The tested FiPy inverse and cloak programs instead use Autograd 1.8
\citep{autograd}, with custom forward and reverse rules registered through
its primitive interface. QuantumOptics uses Julia ChainRulesCore
\citep{julia_chainrules} composed through Zygote \citep{zygote2018}.
Each runtime implements the shared abstraction of a value and derivative
action on declared inputs. Experiment-specific versions and bindings
accompany each program.

For a concrete call, the QuSpin control program loads the package rules and
imports \texttt{vjp} from Python ChainRules. With Hamiltonian object
\texttt{H}, input state \texttt{psi\_in}, fixed duration \texttt{dt} and
solver options, one propagation step is:
\begin{verbatim}
psi_out, pull = vjp(hamiltonian.evolve, H, psi_in, 0.0, dt,
                   wrt=("self", "v0"), **solver_options)
cotangents = pull(output_cotangent)
\end{verbatim}
Here \texttt{self} denotes the Hamiltonian receiver and \texttt{v0} the
initial state. The returned mapping contains their cotangents. Task code
contracts the Hamiltonian cotangent with its control operators and propagates
the state cotangent to the preceding interval. This is the invocation used in
\path{analysis/scripts/run_quspin_robust_control.py} in the supplementary archive. Appendix~\ref{app:quantum-control} gives the control matrices,
objective, solver settings and complex-gradient convention; the reproduction
guide links the task to its required package and source versions.

\paragraph{Separating a solve rule from a research loss.}
\label{app:implicit-solve-rule}
For a real nonsingular system $A(\theta)y=b(\theta)$, an implicit JVP
solves $A\,\delta y=\delta b-(\delta A)y$ \citep{blondel2022}.
For a scalar loss $L(y,\theta)$, reverse differentiation solves
$A^\top\lambda=\partial L/\partial y$.
The solve rule contributes $\lambda^\top[\delta b-(\delta A)y]$ to the
directional derivative of the loss. Task code adds the direct parameter
derivative of $L$. Changing the measured observable changes the loss and
its derivative with respect to $y$, while retaining the solve rule.
This separation lets task code express new research objectives using
the same numerical derivative implementation.

\subsection{Composing a steady-state derivative}
\label{app:task-correspondence}

The QuantumOptics benchmark illustrates how a research objective uses a
shared numerical rule. Task code forms
$H(\theta)=H_0+\sum_k\theta_k C_k$ in a 12-dimensional Hilbert space and
evaluates $\operatorname{Re}\operatorname{tr}(O\rho_{\rm ss}(\theta))$
with a fixed jump operator and rate. The package's
\texttt{steady\_observation} calls the original steady-state solver and
supplies a trace-constrained bordered-Liouvillian adjoint.
Task-level reverse mode connects that rule to the Hamiltonian controls
and observable trace. The benchmark evaluates full-gradient requests at
$P=1,4,16,64,144$; Table~\ref{tab:portfolio-workloads} reports the
144-control endpoint. This tests the composed gradient rather than a
completed inverse-design optimization.

\section{Evaluated capabilities, numerical validation and cost}
\label{app:sources}

The portfolio includes all 20 packages in the assembled gradient-benchmark
collection. Its workloads exercise delivered derivative paths; it is not a
sample of all attempted software-development runs.
Table~\ref{tab:capability-overview-20} identifies representative parameter-to-output
paths and their numerical checks; the task families below describe the broader
scientific uses, complementing the reuse examples in
Table~\ref{tab:capability-reuse}. The research applications
use some of these paths in larger computations; the gradient-cost comparison
in Figure~\ref{fig:portfolio} instead uses one specified workload per package.

\begin{table*}[!htbp]
\caption{Evaluated parameter-to-output capabilities across 20 scientific software packages. Each row identifies the tested numerical boundary, its use of the original computation and its validation. These are evaluated paths, not complete API inventories. Citations identify the original software.}
\label{tab:capability-overview-20}
\centering
\footnotesize
\setlength{\tabcolsep}{3pt}
\renewcommand{\arraystretch}{1.12}
\begin{tabularx}{\textwidth}{@{}>{\raggedright\arraybackslash}p{0.83in}
  >{\raggedright\arraybackslash}X
  >{\raggedright\arraybackslash}p{2.00in}@{}}
\toprule
Software & Evaluated path and original computation & Numerical check\\
\midrule
FiPy & Spatial diffusivities and boundary inputs $\to$ native diffusion fields through implicit solve rules reused in inverse and cloak tasks & Native fields and FD gradients; boundary-map source tests\\
OpenPNM & Throat conductances $\to$ steady Stokes-flow pressure on a fixed pore network & Recorded solver-output and selected-coordinate FD checks\\
PHREEQC \citep{parkhurst2013phreeqc} & Solution inputs $\to$ weighted ionic strength, molalities and log activities & Real-engine speciation and all-input FD checks\\
QuantumOptics & Hermitian controls $\to$ observable of the original steady-state solver through a shared adjoint & Native steady solver and full-gradient FD check\\
QuSpin & Static Hamiltonian parameters $\to$ evolved state, control loss and observables through shared propagation rules & Independent propagation and FD checks; control evaluations\\
BioDynaMo & Voxel concentrations $\to$ one diffusion update through a reimplemented field kernel & Kernel/native-output parity and selected-coordinate FD\\
PhysiCell & Substrate concentrations $\to$ one transport update through a reimplemented two-substrate kernel & Kernel/native-output parity and selected-coordinate FD\\
EvoEF2 & Atomic coordinates of a fixed structure $\to$ molecular stability score & Recorded score and selected-coordinate FD check\\
FreeSASA \citep{mitternacht2016freesasa} & Atomic radii $\to$ total solvent-accessible area on a fixed occlusion branch & Original area calculation and selected-coordinate FD\\
MCell & One rate amplitude $\to$ expected terminal product yield via a model-specific VJP & Analytic reference and native-ensemble FD; seven screened requests\\
REBOUND \citep{rein2012rebound} & Initial masses, positions and velocities $\to$ native many-body energy & Native energy and full-coordinate FD check\\
Cantera & Reaction multipliers $\to$ fixed-state species production rates & Original kinetics state and selected-coordinate FD\\
Kwant & Gate potentials $\to$ native two-lead transmission through an implicit scattering rule; separately, a spectral window primitive & Native transmission, directional FD and VJP pairing; saved moment FD\\
LALSuite \citep{lalsuite2018} & Named parameters $\to$ analytic likelihood with no waveform data & Public analytic callable and all-input FD check\\
ParFlow \citep{kuffour2020parflow} & Pressure samples $\to$ Van Genuchten saturation and relative permeability & Constitutive output and selected-coordinate FD\\
Phonopy \citep{togo2023implementation,togo2023phonopy} & Ambient force constants $\to$ fixed-$q$ dynamical-matrix functional & Recorded selected-coordinate FD; no symmetry constraint\\
SasView \citep{sasview_project} & Wave-vector samples $\to$ fixed-radius sphere form-factor output & Original special-function primal and selected-coordinate FD\\
SHTns & Packed spectral coefficients $\to$ synthesized-field functional & Original transform and selected real-coordinate FD\\
StochKit2 & Rate $\to$ fixed-grid terminal-population PMF and likelihood-ratio derivative & Native SSA and independent one-hot reconstruction\\
TeNPy \citep{hauschild2018tenpy} & Positive-vector entries $\to$ entropy in ambient coordinates & Original entropy primal and selected-coordinate FD\\
\bottomrule
\end{tabularx}
\par\smallskip\parbox{\textwidth}{\footnotesize MCell uses a model-specific expectation rule for one rate amplitude, checked against native Monte Carlo finite differences under a 10\% gradient-error screen.}
\end{table*}

\subsection{Broader workflow capabilities and rule reuse}
\label{app:broader-capabilities}

\paragraph{Transport and fields.}
The OpenPNM workflow report covers pressure-driven flow, molecular diffusion,
heat transport, reaction--diffusion and advection--diffusion. The common
transport and rate interfaces support different coefficients, sources and
flux objectives; Appendix~\ref{app:workflow-expansion} details the reported
source and directed-flow extensions. FiPy's implicit diffusion rules connect
conductivity to temperature fields in both the retained inverse problem and
cloak design. Task code changes the observations and loss while reusing
the solve derivative.

For ParFlow, the report records 80 specified parameter-sensitivity cases.
The inspected simulation interface configures an original \texttt{Run},
invokes the official binary and reads the simulated trajectory. A sidecar
discrete tangent or adjoint propagates sensitivities of the supported
residuals. A workflow test composes this interface with a head/storage
observable and compares its gradient with perturbed original-simulator
runs. The package's saved stage record reports two 349-test runs and a
successful workflow execution. These records support simulator-connected
paths, unlike the pointwise constitutive calculation used in the cost plot;
they are not 80 completed hydrologic inverse problems.

\paragraph{Quantum observables and compositions.}
The workflow evaluations cover more numerical paths than the one-request
cost comparison. QuSpin's static propagation rules can be composed over
piecewise control intervals, as in Section~\ref{sec:quantum-control}.
Kwant provides gate-to-transmission derivatives for the filter design in
Section~\ref{sec:transport-design} and a separate local-density-of-states
derivative for the inverse problem in Appendix~\ref{app:inverse}.
For TeNPy \citep{hauschild2018tenpy}, the report records 42 local AD cases involving
matrix-product states, finite dynamics and related responses. Fixed-source
inspection shows that the same time-evolution interface is tested with
Loschmidt overlap, local and two-point observables, and bond entropy.
The implementation invokes original TeNPy computations for the values,
while several derivative paths use finite-chain or band-response formulas
under explicit domain and value-agreement checks. These bounded routes do
not establish differentiation through arbitrary tensor-network algorithms.
The source tests define how the shared interface is exercised; the reported
42-case execution is a separate workflow-evaluation record.

QuantumOptics provides a source-level example of one rule serving different
objectives: its steady-observation rule is called for both a single
observable and an average over six Hamiltonian offsets. A correlation
objective uses a different rule. Only the single steady scenario is bound to
the retained numerical benchmark record in this study. The other source
definitions identify reusable compositions without supplying additional
measured outcomes. Appendix~\ref{app:task-correspondence} specifies the
measured calculation. The code and data guide records the fixed versions
and separates program definitions, package-reported tests and workflow
evaluation results.

The workflow evaluation separately reports 44 passing cases, including
14 task-level compositions of existing evolution, steady-state and spectral
rules. These evaluated compositions support the QuantumOptics row of
Table~\ref{tab:capability-reuse}; Appendix~\ref{app:workflow-expansion}
describes their evaluation and expansion.

\paragraph{Molecular and cellular computations.}
EvoEF2 \citep{huang2020evoef2} supports binding and stability sensitivities,
including continuous scoring-weight derivatives after mutation, repair or
hydrogen preparation has selected a fixed structure. FreeSASA
\citep{mitternacht2016freesasa} supplies local atomic-radius derivatives of
solvent-accessible area on a fixed visibility branch. These derivatives do
not differentiate the discrete preparation or visibility choices.
BioDynaMo and PhysiCell \citep{breitwieser2022biodynamo,ghaffarizadeh2018physicell}
include field updates and fixed-population responses. The reported
BioDynaMo mechanics rule preserves the original storage and update order;
the timed diffusion kernels are separately checked against original outputs.
These continuous paths exclude population-changing events and discrete
epidemic scheduling.

\paragraph{Chemical and stochastic responses.}
Cantera's evaluated paths \citep{cantera_project} concern thermodynamic,
transport and reaction-rate responses at fixed state. PHREEQC
\citep{parkhurst2013phreeqc} has reported calcium transport and kinetic
calcite-response rules, alongside the separate speciation-gradient demo.
StochKit2 differentiates specified terminal-population statistics, with its
fixed-bin semantics checked in Appendix~\ref{app:stochastic-followup}.
The broader MCell report uses trajectory score-function checks; its
accuracy-screened cost comparison instead uses the model-specific expected
yield rule of Appendix~\ref{app:mcell-rate-amplitude}. These two MCell
results do not share an accuracy guarantee.

\paragraph{Spectral, scattering and gravitational responses.}
Phonopy's workflow evaluations \citep{togo2023implementation,togo2023phonopy}
include force-constant and phonon-frequency responses. SasView
\citep{sasview_project} differentiates scattering-model parameters through
original curve evaluation, with resolution fixed at construction.
SHTns \citep{schaeffer2013shtns} supplies synthesis derivatives
on a fixed declared grid. The broader REBOUND report
\citep{rein2012rebound} covers continuous gravitational-dynamics sensitivities,
whereas its timed request differentiates initial energy. LALSuite's checked
waveform cases \citep{lalsuite2018} concern distance and reference phase for
specified waveform models.
The source-indexed workflow reports support these task families; the
single-request cost plot measures a different slice of their use.

\paragraph{Shared interfaces in recorded executions.}
The routine-case collection contains 1,372 case entries, of which the source
reports mark 938 as passing. Of these, 916 have paired call timings and 22
are classified as forward-only, with no captured derivative calls.
Table~\ref{tab:recorded-call-reuse} identifies
shared call sites in those executions. Counts deduplicate repeated runs and
multiple calls within a case; a case can use several interfaces. They measure
recorded interface use, not independent research discoveries or coverage of
all original-software APIs.

\begin{table}[!htbp]
\centering
\small
\caption{Shared interfaces appearing in routine-case execution records.
The last two columns count distinct case identifiers with any captured call
and with a paired AD/FD timing for that interface, respectively. Repeated
calls and runs count once. Counts do not certify the full scientific
workflow or identify different underlying derivative rules.}
\label{tab:recorded-call-reuse}
\begin{tabularx}{\linewidth}{@{}lXrr@{}}
\toprule
Software & Shared recorded interface & Cases & Paired\\
\midrule
QuSpin & \path{hamiltonian.evolve} & 10 & 2\\
OpenPNM & Transport \path{run} and flux \path{rate} (each) & 66 & 66\\
TeNPy & \path{dmrg_spectral_observable} & 9 & 9\\
TeNPy & \path{tebd_spectral_observable} & 5 & 5\\
Phonopy & \path{homogeneous_frequencies} & 4 & 4\\
LALSuite & \path{SimInspiralChooseFDWaveform} & 16 & 16\\
ParFlow & \path{SimulationResult.jvp} & 80 & 80\\
SHTns & \path{synth_1} & 80 & 80\\
\bottomrule
\end{tabularx}
\end{table}

For QuSpin, ten cases invoke the evolution interface, but only two have a
paired timing for that call; the other eight have AD-only records. OpenPNM
transport and rate calls each occur in 66 cases, although only 22 cases have
paired timing for every captured call. Thus interface-use counts and
complete timing capture answer different questions. Across the collection,
854 cases have fully paired captured calls and 62 have partial capture.
The generic Python probe times perturbed function evaluations without
uniformly assembling and checking FD derivatives. Scientific checks remain
those of the underlying workflow reports; the code and data include separate
call-profile summaries rather than an end-to-end speedup aggregate.

\subsection{Scalar objectives and differentiation coordinates}

Each gradient workload fixes a scalar objective and a set of active numerical
inputs. Table~\ref{tab:benchmark-objectives} summarizes these objectives and
the FD coordinate selections; Table~\ref{tab:portfolio-workloads} reports
their dimensions and costs. Fixed weights, states, geometries and observation
operators are not differentiated unless included among the active inputs.
The code and data guide maps each row to its source function, base input,
fixed weights and seeded coordinate selection; the table is not a replacement
for those executable definitions.

For QuSpin, the real symmetric Hamiltonian has 528 independent coefficients,
rather than 1,024 independent matrix entries. SHTns has 1,024 real degrees of
freedom in packed complex coefficients, with real $m=0$ entries. Phonopy's
coordinate perturbations act in the ambient force-constant array, not a
symmetry- or acoustic-sum-rule-preserving parameterization. TeNPy perturbs
positive vector entries without renormalizing them onto a probability simplex.
These choices define the derivatives being compared.

\begingroup
\footnotesize
\setlength{\tabcolsep}{3pt}
\renewcommand{\arraystretch}{1.15}
\begin{longtable}{@{}>{\raggedright\arraybackslash}p{0.14\linewidth}
 >{\raggedright\arraybackslash}p{0.53\linewidth}
 >{\raggedright\arraybackslash}p{0.29\linewidth}@{}}
\caption{Summary of scalar objectives and active inputs for the 20 workloads in
Figure~\ref{fig:portfolio}. Weights, observation operators and stated model
conditions are fixed; the last column identifies the FD request.
Timing and component counts appear in Table~\ref{tab:portfolio-workloads}.}
\label{tab:benchmark-objectives}\\
\toprule
Software & Scalar objective and active input & Headline FD coordinates \\
\midrule
\endfirsthead
\multicolumn{3}{l}{Table \thetable\ (continued)}\\
\toprule
Software & Scalar objective and active input & Headline FD coordinates \\
\midrule
\endhead
\bottomrule
\endfoot
FiPy & Weighted steady diffusion--reaction solution; 808 face diffusivities on fixed $24\times16$ grid. & Seeded 256 faces. \\
OpenPNM & Weighted Stokes-flow pressure; 565 throat conductances on fixed 300-pore network. & Seeded 256 throats. \\
PHREEQC & Weighted ionic strength, molalities and log activities; six seawater totals plus pH, pe, alkalinity, temperature and pressure. & All 11 named inputs. \\
QuantumOptics & $\operatorname{Re}\operatorname{tr}(O\rho_{\rm ss})$ with $O=n/11+0.08(a+a^\dagger)$ and a controlled 12-dimensional Hamiltonian; 144 real Hermitian-basis controls. & All 144 active controls; base and active set vary across the smaller-$P$ series. \\
QuSpin & Real target-state overlap after $32\times32$ symmetric-Hamiltonian evolution; 528 independent real matrix coefficients. & First 256 symmetric directions. \\
BioDynaMo & Weighted output of one periodic $16^3$ diffusion step; 4,096 input concentrations in an extension kernel. & Seeded 256 voxels. \\
PhysiCell & Weighted output of one $32^2$, two-substrate diffusion--decay step; 2,048 input concentrations in an extension kernel. & Seeded 256 substrate voxels. \\
EvoEF2 & Total stability score of fixed 968-atom 1aac structure; 2,904 Cartesian coordinates. & Seeded 1,024 coordinates. \\
FreeSASA & Total Shrake--Rupley accessible area of fixed 602-atom 1ubq; atomic radii on a fixed occlusion-mask branch. & First 256 radii. \\
MCell & Expected terminal B count for 100 initial A molecules; one amplitude $\alpha$ scales 128 first-order A$\to$B rates. & One amplitude direction ($P=D=1$), centered step 0.2 with native paired-seed ensembles. \\
REBOUND & Native total initial-state energy of 32 bodies; 32 masses, 96 positions and 96 velocities. & All 224 coordinates in seeded order. \\
Cantera & Weighted 53-species net-production rates at fixed $T$, $P$ and composition; 325 reaction multipliers. & First 256 multipliers. \\
Kwant & Weighted Jackson-window output; 4,096 synthetic Chebyshev moments, without a Hamiltonian-to-moments chain. & Seeded 256 moments. \\
LALSuite & Correlated analytic likelihood with data and model arguments set to \texttt{None}; 15 CBC-named inputs. & All 15 named inputs. \\
ParFlow & Weighted Van Genuchten saturation and relative permeability; 4,096 negative pressure-head values with constitutive parameters fixed. & Seeded 256 cells. \\
Phonopy & Real contraction of fixed-$q$ 8-atom dynamical matrix; 576 ambient real force-constant entries. Perturbations need not preserve symmetry or acoustic sum rules. & Seeded 256 entries. \\
SasView & Weighted squared sphere form-factor amplitudes at fixed radius; 4,096 independent $q$ locations. & Seeded 256 $q$ samples. \\
SHTns & Real contraction of a $64\times96$ synthesized field; 1,024 real degrees in packed complex coefficients. & First 256 packed real degrees. \\
StochKit2 & Weighted initial-time propensities of 64 independent decay channels; rates at fixed initial populations. & All 64 rates; initial-time values are deterministic. \\
TeNPy & Entropy of a synthetic positive 1,024-entry vector; ambient entries, with no FD renormalization. & Seeded 256 entries. \\
\end{longtable}
\endgroup

\subsection{Numerical checks}
\label{app:costs}
\label{app:normalized-cost}

Forward-value agreement, derivative accuracy and attainment of a research
objective are checked separately. JVP/VJP duality tests internal consistency;
FD, analytic derivatives or independent numerical implementations check the
intended derivative. For the selected headline coordinates, the recorded
maximum scaled differences
$\max_i |g_i^{\mathrm{AD}}-g_i^{\mathrm{FD}}|/\max(1,|g_i^{\mathrm{FD}}|)$
are $3.81\times10^{-10}$ (FiPy), $5.12\times10^{-8}$ (EvoEF2) and
$5.09\times10^{-7}$ (PHREEQC). QuantumOptics reports a relative gradient
$L_2$ difference of $1.16\times10^{-8}$. These metrics check the defined
callables and are not interchangeable measures of whole-package accuracy.

\begin{figure}[!htbp]
    \centering
    \includegraphics[width=\linewidth]{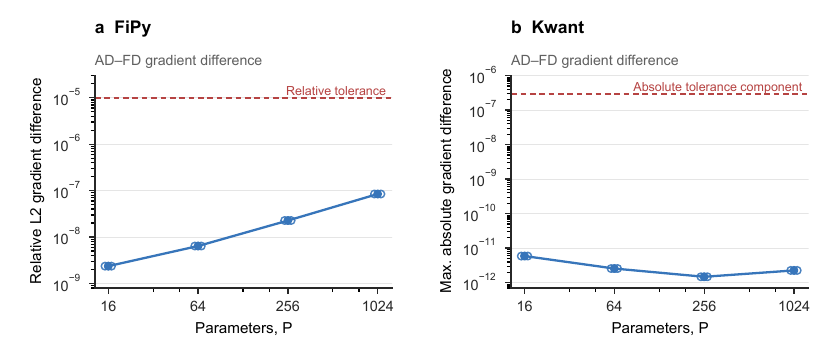}
    \caption{AD/FD numerical checks for the workloads in
        Figure~\ref{fig:gradient-scaling}.
        (a) FiPy relative $L_2$ gradient differences.
        (b) Kwant maximum absolute component differences.
        Points show all three process records per size; lines connect their medians.
        Dashed lines mark the configured relative tolerance for FiPy and the absolute
        component of Kwant's combined absolute/relative tolerance. Differences from
        FD are numerical comparisons, not errors against an exact derivative.}
    \label{fig:gradient-validation}
\end{figure}

Figure~\ref{fig:gradient-validation} gives all 24 process/size checks for the
separate FiPy and Kwant full-gradient series. The FiPy relative $L_2$
differences are cross-checked against stored arrays. Kwant's maximum absolute
difference is recomputed from the stored gradients; its comparison uses
absolute tolerance $3\times10^{-7}$ and relative tolerance $4\times10^{-5}$.

\subsection{Twenty-package gradient requests and timings}
\label{app:additional-profiling}

Table~\ref{tab:portfolio-workloads} gives the requests and measured costs in
Figure~\ref{fig:portfolio}. For coordinate-request scans, the figure uses the
largest measured request, not the largest observed speed ratio. Fourteen rows
compare selected-coordinate FD with an AD call returning the complete
gradient. Selecting those AD components for the error check occurs after
timing; no full-FD cost is extrapolated. The remaining rows request all
task-level coordinates, including MCell's single rate-amplitude derivative.

\begin{table}[p]
\centering
\caption{Definitions and reported timing medians for all workloads in Figure~\ref{fig:portfolio}.
$P/D$ gives derivatives requested by FD / the complete task-level gradient returned by AD (after contraction for MCell); when $P<D$, FD requests a selected subset contained in the full AD result, not the same complete gradient.
$n$ is the source-reported repeat count, not independent processes. Times are milliseconds and follow the Figure~\ref{fig:portfolio} ratio order. MCell ($P=D=1$) uses five $\alpha=1$ requests for timing medians; two $\alpha=0.8$ requests validate accuracy and are excluded from those medians.}
\label{tab:portfolio-workloads}
\footnotesize
\setlength{\tabcolsep}{3pt}
\renewcommand{\arraystretch}{1.2}
\begin{tabularx}{\linewidth}{@{}lXrrrr@{}}
\toprule
Software & Computation & $P/D$ & $n$ & FD (ms) & AD (ms) \\
\midrule
FiPy & 24$\times$16 diffusion-reaction solution & 256/808 & 7 & 1,670 & 10.8 \\
OpenPNM & 300-pore steady Stokes-flow pressure & 256/565 & 9 & 419 & 2.22 \\
PHREEQC & seawater equilibrium observables & 11/11 & 5 & 58.9 & 3.21 \\
QuantumOptics & 12D steady-state observable & 144/144 & 3 & 4,570 & 16.5 \\
QuSpin & 32$\times$32 symmetric-Hamiltonian evolution & 256/528 & 5 & 74.5 & 0.592 \\
BioDynaMo & extension-kernel 16$^3$ diffusion step & 256/4096 & 9 & 20 & 0.12 \\
PhysiCell & extension-kernel 32$^2$ two-substrate step & 256/2048 & 7 & 729 & 4.26 \\
EvoEF2 & 968-atom stability score & 1024/2904 & 5 & 3,320 & 49.8 \\
FreeSASA & 602-atom accessible area & 256/602 & 5 & 587 & 2.62 \\
MCell & expected reaction yield: rate-amplitude sensitivity & 1/1 & 5 & 71,700 & 39 \\
REBOUND & 32-body initial energy & 224/224 & 7 & 85.6 & 1.54 \\
Cantera & 53-species production rate; 325 reaction multipliers & 256/325 & 7 & 15.1 & 0.477 \\
Kwant & Chebyshev window functional & 256/4096 & 9 & 19.5 & 0.0817 \\
LALSuite & 15-parameter analytic likelihood & 15/15 & 51 & 0.158 & 0.0183 \\
ParFlow & 64$^2$ soil response & 256/4096 & 9 & 45.6 & 0.589 \\
Phonopy & 8-atom dynamical-matrix functional & 256/576 & 9 & 268 & 1.96 \\
SasView & sphere form-factor q-sample sensitivity & 256/4096 & 11 & 27.3 & 0.128 \\
SHTns & 64$\times$96 spherical synthesis functional & 256/1024 & 7 & 7.58 & 0.139 \\
StochKit2 & 64-channel propensity sum & 64/64 & 5 & 1,340 & 9.36 \\
TeNPy & 1024-entry entropy & 256/1024 & 9 & 2.82 & 0.0321 \\
\bottomrule
\end{tabularx}
\par\smallskip\parbox{\linewidth}{\footnotesize For rows other than MCell, the listed FD request is the largest coordinate request in its source series, not a maximum-speedup selection.
QuSpin's 1,024 stored matrix entries map to 528 independent real controls; SHTns has 1,024 independent real coordinates.
BioDynaMo and PhysiCell use reimplemented diffusion steps checked against original-software outputs; timings are for these kernels.
MCell compares a model-specific analytical expectation rule with native Monte Carlo FD under seven AD-reference and FD 10\% accuracy checks. Its bar divides median complete FD request time by median public AD VJP/contraction time. The public 128-rate VJP is contracted to one $\alpha$ derivative; no 128-coordinate acceleration is implied.
QuantumOptics and MCell supply individual within-process timing repeats; no confidence intervals are inferred for the other rows.}
\end{table}

Most request scans keep the physical problem fixed. QuantumOptics instead
changes the active Hamiltonian-control set on a fixed 12-dimensional Hilbert
space; at 144 controls, both methods return the full gradient and FD uses
$2P+1$ steady-state solves. It supplies three within-process timing samples.
MCell supplies five primary requests and two additional accuracy checks.
For the other 18 workloads, median timings are available but the individual
repeats are not retained, so no between-run uncertainty is inferred.

Two rows, BioDynaMo and PhysiCell, apply AD and FD to the same reimplemented
diffusion kernels rather than time native whole-simulator execution.
Separate tests check selected BioDynaMo stencil outputs and a complete
PhysiCell transport-step field against the original implementations, using
inputs different from the timing demos. FD agreement alone does not establish
this original-software fidelity.

StochKit2's headline request concerns deterministic initial propensities at
fixed populations, not a derivative through stochastic evolution. FD makes
128 objective calls for 64 rates and returns only the gradient; AD returns
value and gradient. Both clear the run cache for each timed sample. Its
143-fold ratio includes both complete calling paths, without isolated primal
timings to explain their per-call difference. MCell instead compares a
model-specific expectation rule with sampled native FD at a specified
accuracy (Appendix~\ref{app:mcell-rate-amplitude}); its savings include avoided
sampling, not reuse across many gradient components.

PHREEQC and EvoEF2 likewise time FD gradients without an unperturbed
value, using 22 and 2,048 objective calls, respectively; their AD
requests return both value and gradient.

Some source summaries identify a checkout but not the modules imported by
the timed process. Those records support the reported workload costs, not
a verified comparison between software revisions. The controlled
implementation comparisons bind both executed versions and use matched
inputs and environments.

Outside this headline set, a separate 65-momentum Kwant band-value/JVP
request remains slower than FD after preparation reuse (2.462 versus
1.657 ms). Its definition and version-comparison records accompany the
code and data; it is not the KPM workload in Figure~\ref{fig:portfolio}.
The ten-task FiPy comparison in Appendix~\ref{app:implementation-changes}
also includes requests where the later AD version remains slower than FD.
These request-level costs differ from the full optimization intervals in
Appendix~\ref{app:inverse} and the matched implementation retests in
Appendix~\ref{app:software-revisions}.

\subsection{Complete-gradient scaling and computational work}
\label{app:gradient-scaling}

\begin{figure}[t]
    \centering
    \includegraphics[width=\linewidth]{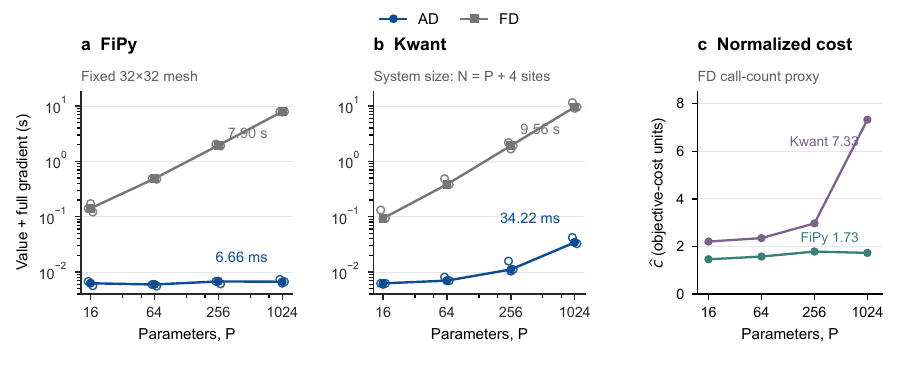}
    \caption{Complete-gradient cost as the parameter dimension grows.
    (a,b) Value plus full gradient: FiPy fixes its $32\times32$ mesh; Kwant grows
    to $N=P+4$ sites at energy 0.23. Open points are within-process medians of
    three timings; lines connect medians over three processes.
    (c) FD-normalized AD cost, $\widehat c=(2P+1)t_{\rm AD}/t_{\rm FD}$,
    using method-specific medians, without process pairing.
    Each measurement is one value-and-full-gradient request.
    Derivative checks are in Figure~\ref{fig:gradient-validation}.}
    \label{fig:gradient-scaling}
\end{figure}

Central FD repeats a forward calculation for each active parameter.
The complete-gradient series in Figure~\ref{fig:gradient-scaling} grows
to 1,024 parameters: FiPy takes 6.66 ms with AD versus 7.90 s with FD;
Kwant takes 34.2 ms versus 9.56 s. FiPy keeps its $32\times32$ physical mesh
fixed, whereas Kwant grows the system to $N=P+4$ sites at energy 0.23.
Each method has three separate processes and three timing samples per
process at each size.

To express the value-and-gradient cost in forward-objective units, define
\begin{equation}
    \widehat c=N_{\mathrm{FD}}\frac{t_{\mathrm{AD}}}{t_{\mathrm{FD}}},
    \label{eq:cost}
\end{equation}
where $N_{\mathrm{FD}}=2P+1$ includes the unperturbed value in these two
series. This proxy assumes similar costs among the timed FD objective calls;
it includes assembly and wrapper work and is not an isolated reverse-pass
measurement. Using method-specific medians without process pairing,
$\widehat c$ ranges from 1.46 to 1.78 for FiPy and from 2.21 to 7.33 for
Kwant. The latter trend reflects changing system size as well as parameter
count and does not by itself identify the cause of the growth.

The Kwant instrumentation records two builds, two factorizations and three
solves per AD request, versus $2P+1$ builds/factorizations and $2(2P+1)$
solves for FD. Constant counts do not imply constant cost when each linear
system grows. Complete-optimization ratios additionally depend on the
number of evaluations and on the number of loads or energies per objective;
Appendix~\ref{app:inverse} gives those separate comparisons.

\subsection{MCell: expected-yield sensitivity at a fixed accuracy target}
\label{app:mcell-rate-amplitude}

\paragraph{Task and derivative.}
We use native MCell 4.1.0 \citep{husar2024mcell} to simulate irreversible
A$\to$B conversion of 100 initially released molecules over 128 intervals,
each of duration $\Delta t=10^{-6}$\,s. One parameter $\alpha$ scales the
prescribed rate profile, $k_j(\alpha)=\alpha k_j^0$, with
$k_j^0=8000+4000j/127$\,s$^{-1}$. The target is the sensitivity of expected
final yield, not the derivative of a fixed-seed integer count:
\[
    F(\alpha)=\mathbb E[B(T)]=100(1-e^{-\alpha S_0}),\qquad
    q(\alpha)=100S_0e^{-\alpha S_0},\qquad
    S_0=\Delta t\sum_j k_j^0=1.28.
\]
The package at revision \texttt{c5be7307} provides a model-specific analytical
expectation rule alongside native stochastic forward values. One public
VJP returns the rate derivatives; contracting them with the base rates
$k_j^0$ gives $q(\alpha)$. The benchmark adds only this parameterization and
contraction, not a replacement derivative implementation. Both methods
therefore target the same one-parameter expectation gradient ($P=D=1$).

\paragraph{Precision and complete requests.}
FD uses an absolute step $h=0.2$ and 2,048 native trajectories per side,
organized as 32 independent blocks of 64 seeds. Each plus/minus pair
shares seeds. The step has approximately 1.10\% analytical truncation bias,
which is retained in the raw estimate rather than corrected using the
reference. Each request must satisfy
\[
    \frac{|\bar q_{\rm FD}-q|+t_{0.975,31}\operatorname{SE}_{32}}{|q|}
    \leq 0.10,\qquad t_{0.975,31}=2.0395.
\]
These are nominal block-based intervals for individual requests, not a
simultaneous coverage guarantee. The public AD contraction is independently
checked to relative tolerance $10^{-8}$. Its two ancillary native forward
trajectories, pullback and contraction are timed; FD timing includes all
64 public ensemble calls and estimate assembly. Base model construction,
verification, warmup and separate forward-value checks are outside these
gradient-only request timings.

The protocol fixes five primary requests at $\alpha=1$ and two validation
requests at $\alpha=0.8$, with fresh, disjoint seed families.
All seven pass: the largest recorded AD relative error is
$2.0\times10^{-16}$, and the FD error-screen values span 1.52--3.66\%.
Separate 2,048-trajectory native means are 72.272 and 64.103, consistent
with the model expectations 72.196 and 64.084 within their nominal 95\%
block intervals.

\paragraph{Observed costs and interpretation.}
The five primary requests have median AD and FD times of 0.03896 and
71.6914\,s, respectively. Figure~\ref{fig:portfolio} uses their ratio,
approximately $1840\times$. The comparison is between a one-parameter
analytical expectation rule and native Monte Carlo FD at the stated precision.
The two validation AD calls take 2.214 and 0.0386\,s; both are retained.
Only the five predeclared primary requests enter the timing medians.

Execution uses Python 3.12.13, NumPy 2.5.1 and ChainRules 0.1.0 on a
16-CPU shared development host. Independent reanalysis checks seed assignments, native counts, AD contractions
and FD statistics. The protocol, runner and raw records accompany the results.

\section{Research-driven software revisions}
\label{app:software-revisions}

The cases in Section~\ref{sec:recursive-results} and
Table~\ref{tab:evolution-cases} connect research findings to changed
implementations or compositions. Kwant and StochKit2 are supported by
direct executions before and after revision. Additional workflow reports
specify tested software versions, inputs, observables and numerical checks.
The sections below give these comparisons and their source correspondence;
case counts describe tested tasks, not fractions of a package's complete API.

\begin{table}[!htbp]
\caption{Research requirements, retained software changes and subsequent uses.
QuSpin's migrations and the OpenPNM, BioDynaMo and QuantumOptics outcomes are
reported workflow retests; later QuSpin control directly uses the package rule.
The surrounding sections give source identities and checks.}
\label{tab:evolution-cases}
\centering
\small
\setlength{\tabcolsep}{2.5pt}
\renewcommand{\arraystretch}{1.08}
\begin{tabularx}{\linewidth}{@{}YYY@{}}
\toprule
Research requirement & Retained response & Subsequent use or checked result\\
\midrule
Kwant: tune gates to filter electron energies
& Scattering derivative, then shared linearization
& Filter targets met; saved device reused to verify matching gradients at lower cost (Fig.~\ref{fig:recursive-evidence})\\
StochKit2: differentiate the requested population bins
& Fixed-bin rules and reusable simulated trajectories
& Numerical reconstruction agrees; later reuse preserves outputs with calls $4\to2$\\
OpenPNM: include reaction sources and directed transport
& Source-term, directed-conductance and flux rules in the report-tested patch
& Reaction--diffusion and advection--diffusion join the checked paths: $42\to66$ of 80 pass\\
BioDynaMo: preserve the original mechanics updates
& Rule matching sequential updates and storage reorder
& Eleven partial mechanics cases pass after repair\\
QuSpin: use shared propagation derivatives in research programs
& Package rule differentiating Hamiltonian-active evolution
& Eight programs migrated; two more pass ($23\to25$/47); later control reuses the rule\\
\midrule
QuantumOptics: compose further inference and control tasks
& Nine programs using existing evolution, steady and spectral rules
& Workflow retests: $35\to44$ of 49 pass; no new kernel implied\\
\bottomrule
\end{tabularx}
\par\smallskip
\raggedright OpenPNM's count belongs to the report-tested patch; the corresponding
retained commit has not been shown to be byte-identical to that execution
(Appendix~\ref{app:workflow-expansion}).
\end{table}

\subsection{Kwant transport extension and filter design}
\label{app:transport-design}

\paragraph{From a research requirement to a shared rule.}
The original task used Kwant's public scattering solver to evaluate a
two-terminal device, but package revision \texttt{9d911433} did not expose the
required gate-parameter derivative through that solver. Research feedback
identified this missing path. Revision \texttt{d654f20a} added
\texttt{smatrix\_solve\_result}: the original Kwant scattering call supplies
the forward value, and an implicit derivative of its augmented linear system
supplies transmission JVPs and VJPs. Leads, topology and energy are fixed;
the research program supplies local Hamiltonian derivatives, not a replacement
scattering derivative. The code and data records preserve both source revisions, the native
forward calculation and the independent retest.

\paragraph{Device and objective.}
The scattering region is a $24\times6$ square lattice with nearest-neighbor
hopping $-1$ and onsite potential
\begin{equation}
    V_\theta(x,y)=4+\sum_{j=1}^{8}\theta_j
    \exp\!\left[-\frac{(x-x_j)^2}{2(2.0)^2}
        -\frac{(y-y_j)^2}{2(1.25)^2}\right].
\end{equation}
The centers enumerate $x_j\in\{5,9,13,17\}$, with $y_j=1.5,3.5$ at each
$x_j$. Two fixed ideal leads have onsite 4, hopping $-1$, and two open
channels throughout the tested energies. Energies and gate amplitudes are
in units of the hopping magnitude. The six training energies are
$\{0.95,1.00,1.05,1.25,1.30,1.35\}$, with normalized targets
$y=(0,0,0,1,1,1)$. We minimize
\begin{equation}
    L(\theta)=\frac{1}{6}\sum_{i=1}^{6}
    \left[\frac{T(E_i;\theta)}{2}-y_i\right]^2
    +\frac{10^{-3}}{8}\|\theta\|_2^2,
    \qquad \theta\in[-1.5,1.5]^8.
\end{equation}
Both routes start from $(0.18,-0.12,0.10,-0.08,0.08,-0.06,0.04,-0.03)$
and use L-BFGS-B with \texttt{ftol} $10^{-10}$, \texttt{gtol} $10^{-6}$,
at most 40 accepted updates and 20 line-search steps. Each route has a
6,000-scattering-assembly and 300-second budget. Central FD uses step
$10^{-4}$ for each parameter; at a box boundary its stencil may extend
outside the box, while optimizer iterates remain constrained. These settings
and the evaluation rule were fixed before execution.

\paragraph{Independent checks.}
Two parameter profiles at 17 energies retain native forward and flux
conservation checks. Reexpressing the original parameter vector as the
new interface's eight named scalars leaves the Hamiltonian exactly unchanged.
Before optimization, two directions, each perturbing all eight gates, are checked at
energies 0.95 and 1.35 for both profiles. Maximum directional AD/FD differences
decrease from $2.69\times10^{-7}$ to $2.69\times10^{-9}$ and
$6.66\times10^{-11}$ as the step decreases from $10^{-3}$ to $10^{-4}$ and
$10^{-5}$; JVP/VJP pairing differs by at most $2.78\times10^{-15}$.
These tests invoke the public rule and actual native solver.

\paragraph{Design result at reserved energies.}
Evaluation uses four stop-band energies
$\{0.9625,0.9875,1.0125,1.0375\}$ and four pass-band energies
$\{1.2625,1.2875,1.3125,1.3375\}$, all excluded from optimization.
The predeclared criterion is mean stop-band transmission at most 0.5,
mean pass-band transmission at least 1.5, and flux conservation at every
evaluation point. Evaluation always uses the last accepted iterate, not a
point selected using the reserved energies.

\begin{figure}[htbp]
    \centering
    \includegraphics[width=\linewidth]{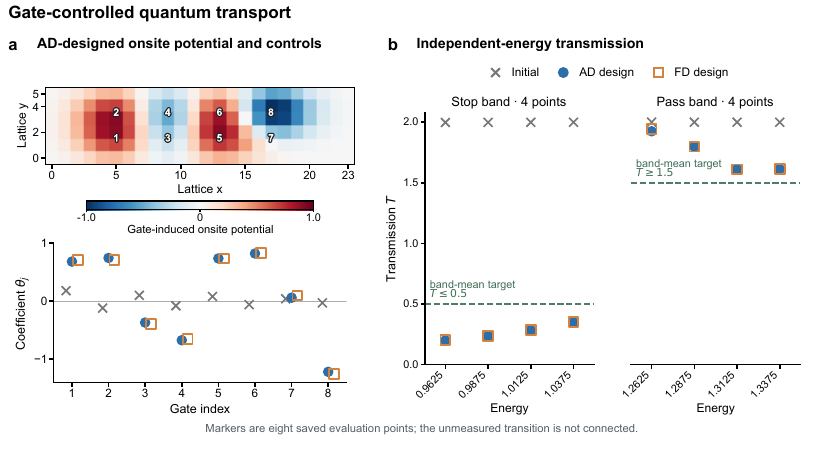}
    \caption{Native transport design after the scattering-rule addition.
        (a) The gate-induced potential reconstructed from the saved AD design and
        the eight coefficients for initial, AD and FD designs. The heatmap excludes
        the constant onsite term 4. (b) All eight reserved-energy transmissions;
        the dashed targets concern band means. Only measured energies are plotted,
        with no interpolation across the transition. Both designs meet the criterion;
        they are the last accepted iterates at the budgets in
        Table~\ref{tab:gate-filter-results}.}
    \label{fig:gate-filter-detail}
\end{figure}

\begin{table}[htbp]
    \caption{Reserved-energy performance of the fixed-device designs. Each
        band mean uses four points. AD stops at its update limit; FD stops at its
        assembly limit and retains the last accepted iterate.}
    \label{tab:gate-filter-results}
    \centering\small
    \begin{tabular}{@{}lrrrr@{}}
        \toprule
        Design      & Stop mean & Pass mean & Accepted updates & Assemblies \\
        \midrule
        Initial     & 1.99856   & 1.99998   & ---              & ---        \\
        Package VJP & 0.26580   & 1.73687   & 40               & 3,078      \\
        Central FD  & 0.26751   & 1.74083   & 38               & 6,000      \\
        \bottomrule
    \end{tabular}
\end{table}

All eight evaluated points also individually meet the numerical band
thresholds, although the preregistered criterion uses band means. Across
the four designs (ungated, initial, AD, FD) and eight energies, the maximum
flux-conservation residual is $4.00\times10^{-15}$. This is a result at
the tested energies, not a continuous-band or disorder-robustness guarantee.
The two final training losses are 0.02082 and 0.02107, versus 0.49921 initially.
The terminal designs exclude the interrupted FD query. Instrumented
wall times are 274.62 and 122.42 seconds.
Different termination conditions and instrumentation prevent interpreting
these as a matched optimization speed ratio.

\paragraph{Dense evaluation of the fixed designs.}
A subsequent forward-only study samples $E=0.900,0.905,\ldots,1.400$ for the
saved initial, AD and FD parameter vectors, with no reoptimization or selection.
All 303 native scattering calls complete with two channels per lead and a
maximum flux residual of $1.35\times10^{-14}$.
At the 21 sampled points in each specified band, both final designs satisfy
$T\leq0.5$ in $[0.95,1.05]$ and $T\geq1.5$ in $[1.25,1.35]$.
The AD stop/pass means are 0.26738/1.73460; FD gives 0.26911/1.73743.
Transmission rises over the 39 sampled transition points, then falls above
the pass band: at $E=1.360$, AD and FD designs give 0.97906 and 1.04149.
Figure~\ref{fig:recursive-evidence}b retains this upper-energy behavior
for the AD-designed device.
The grid characterizes previously evaluated designs and is not a new blind
test or a guarantee between samples. The code and data records contain the settings and all sampled values.

\paragraph{From a computational finding to a second package revision.}
The first VJP constructs its eight components through eight directional
solves, repeating the scattering linearization at a fixed parameter point.
At the final design, a step-refinement check gives approximately quadratic
central-FD error decay, and five-point FD agrees within
$3.50\times10^{-10}$. The version comparison below uses independently
checked derivatives at matched fixed inputs.

Revision \texttt{a93c20b3} subsequently introduces a shared scattering
linearization and a scalar-output adjoint, replacing the repeated directional
reconstruction in \texttt{d654f20a}. We independently retest both exact versions
on the same initial and saved AD-designed device, without reoptimization.
For each six-energy value-and-gradient request, measured system assemblies
fall from 54 to 12 and derivative factorizations from 48 to six; six additional
native primal factorizations remain in each version. Full gradients agree
between versions to at most $7.15\times10^{-16}$. Central FD at step $10^{-5}$
has maximum absolute discrepancy $1.41\times10^{-10}$ at the initial point
and $4.65\times10^{-8}$ at the designed point; two-direction, two-step checks
and JVP/VJP pairings also pass.

Unprofiled measurements use three fresh processes per method, each with one
warmup and one timed call per point. Process order rotates old/new/FD,
new/FD/old and FD/old/new. Initial-point medians are 1.3470/0.4811/0.8801\,s;
designed-point medians are 1.3433/0.4813/0.8821\,s. Thus the revised call costs
about $2.8\times$ less than the earlier implementation and $1.83\times$ less
than same-environment FD in this eight-parameter task. The runtime uses one
thread, Python 3.12.13, NumPy 1.26.4, SciPy 1.13.1 and ChainRules 0.1.0,
on a non-exclusive Xeon virtual machine with the SciPy sparse backend.
Counts are collected separately from timings. This is a bounded paired cost
comparison, not a whole-optimizer speedup or a parameter-scaling study;
parameter-dependent checks and gradient accumulation still incur work.
The code and data materials include both implementations' checks, operation
counts, timings and the replay program.

\subsection{StochKit2: fixed-bin derivatives and simulation reuse}
\label{app:stochastic-followup}

For a terminal-population histogram, the primal and its likelihood-ratio
derivative must use the same bin boundaries. Research use found that
revision \texttt{982237e1} returned the explicitly requested histogram but
recomputed automatic bin bounds inside its derivative. With terminal
populations $[0,2]$, supplied scores $[-1,1]$, and two bins on $(-0.5,5.5)$,
the correct fixed-bin probability vector is $[1,0]$ and its derivative is
$[0,0]$. The old rule instead returned $[-0.5,0.5]$, while still passing
JVP/VJP duality. Revision \texttt{9d098b26} forwards and validates the
specified grid. Independent old/new package executions reproduce this
distinction; a second fixture checks nonzero derivatives and endpoint clipping.
The regressions replace only trajectory generation with fixed traces;
package rules, binning and composed losses execute.

Native SSA checks then use
$\varnothing\xrightarrow{k_1}X$,
$\varnothing\xrightarrow{k_2}2X$ and
$X\xrightarrow{d}\varnothing$, starting at zero with $(k_1,k_2,d)=(6,2,1)$
and terminal time 2 in the simulator's chosen time unit, with rates per
that unit rather than a material-specific physical calibration.
The direction $(-2,1,0)$ leaves the mean unchanged
but changes the population distribution, motivating a distribution-sensitive
objective rather than mean fitting alone. Independent compound-Poisson
probabilities and their derivatives supply a reference; exported sufficient
statistics independently check the package's likelihood-ratio calculation.
The histogram has 51 bins on $(-0.5,50.5)$; out-of-range samples are clipped
to endpoint bins, and none occur in the recorded native ensembles.

All three initial 512-trajectory ensembles pass fixed-bin value/JVP/VJP
and reported-standard-error arithmetic checks. One fails its predeclared
analytic-gradient screen: an empty bin has zero empirical derivative and
standard error despite a nonzero exact derivative. The failed record remains.
A subsequent fixed-protocol study uses four fresh seeds at each of 8,192
and 131,072 trajectories. The relative gradient L2 error of the four-seed
mean falls from 0.732 to 0.199; single-run errors at the larger size remain
0.219--0.532. Four replicates describe precision, not uncertainty-interval
coverage or completed parameter fitting. Per-bin counts and score statistics
from the native drivers are retained for reconstruction.

The fixed-grid repair subsequently exposes duplicate native simulation when
the requested bin count differs from the default. Revision \texttt{6fdbf85}
aligns the value and derivative requests and excludes bin count from trajectory
cache identity when official histogram files are not requested. We independently
compare it with \texttt{9d098b26} on the native pure-death model: initial
population 100, rate one, horizon one, 20,000 realizations, four observation
intervals and fixed seed 20260922. The 51-bin grid is $(-0.5,50.5)$ with
endpoint clipping. This is a different network from the precision study above.

Three fresh-process, empty-package-cache repeats per version alternate
old/new order. Each executes the public value, JVP and VJP sequence. Actual
native calls fall from four to two in every repeat: the earlier version runs
both official SSA and the statistics driver twice, with identical trace hashes;
the repaired version runs each once. All value and derivative arrays are bitwise
identical between versions and match the independently implemented fixed-grid
likelihood-ratio reconstruction executed in the probe. A separate native check
retains distinct eight-bin and sixteen-bin outputs when official histograms
are requested. No distribution fit or statistical-accuracy claim follows.

Instrumented diagnostic medians are 1.215 and 0.636\,s, including per-call
recording and executable hashing; these are not uninstrumented package timings
or an isolated speedup estimate. The retained result is elimination of a
duplicate seeded simulation without changing the estimator. Execution used
Python 3.12.13, NumPy 2.5.1, ChainRules 0.2.0 and the same compiled native
drivers for both versions. The records include both source versions, complete output arrays and
the replay program.

\subsection{FiPy execution changes and additional implementation evidence}
\label{app:implementation-changes}

A FiPy molecular-messenger workflow constructs equation objects 150 times
across forward and derivative rollouts, while making 150 solver steps.
Reusing equation workspaces and field maps reduces construction to one;
structures are rebuilt when they change. The revision also uses reverse-mode
task code for scalar multi-parameter objectives. A matched three-process
comparison gives old/new AD medians of 3.816 and 1.408 seconds, with
frozen-output agreement. This measures the combined package and task-code
revision, not the isolated effect of either change or reuse of a numerical
factorization across parameter values.

\begin{figure}[!htbp]
    \centering
    \includegraphics[width=\linewidth]{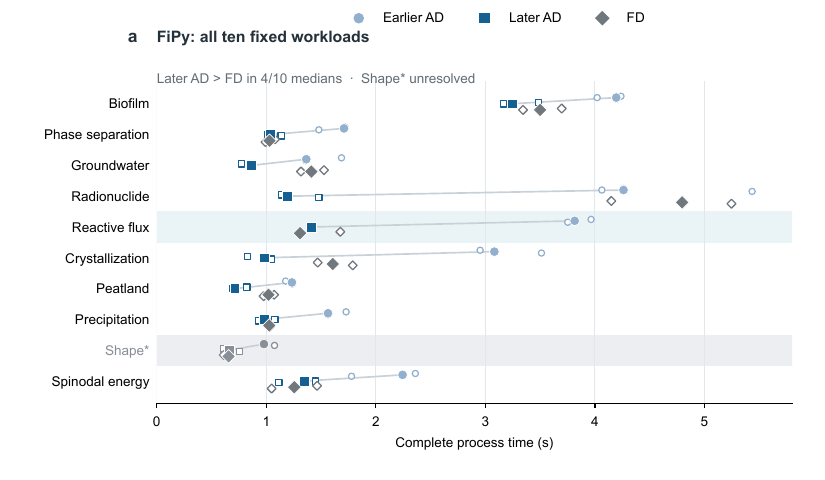}
    \caption{FiPy execution comparison across ten research workloads.
        Open symbols show three separate process measurements per method and workload;
        filled symbols are medians. All ten workloads are retained, including Shape*
        with its unresolved derivative check and the four later-AD medians slower
        than FD. Revisions combine package preparation reuse and reverse-mode task code.}
    \label{fig:fipy-repeats}
\end{figure}

Shape* in Figure~\ref{fig:fipy-repeats} is a level-set control sweep followed
by fixed-branch upwind transport and a smooth objective. Its timing runs
reproduce the saved outputs, but a stricter FD check of the near-zero
derivative remains unresolved. That row supports an execution-cost comparison,
not independently verified derivative accuracy. The workflow definitions
and separate checks are indexed in the code and data guide.

FiPy's separate boundary-vector repair has field/solve source tests on one-cell grids in one,
two and three dimensions; the performance comparison uses two versions
after that repair and does not isolate its effect.

\subsection{Workflow-driven capability repair and expansion}
\label{app:workflow-expansion}

\paragraph{QuSpin: moving propagation derivatives into shared software.}
Revision \texttt{3012e48} adds Hamiltonian-active JVP/VJP rules for
\texttt{hamiltonian.evolve}, calling original QuSpin propagation for values
and using matrix-exponential Fr\'echet derivatives for a static finite
Hamiltonian. The workflow report distinguishes eight previously passing
programs migrated from caller implementations from two newly passing
derivative programs, increasing the evaluated total from 23 to 25 of 47.
The new programs check a dynamics-learning Jacobian and a gate-synthesis
directional derivative; they are not completed learning or design runs.
Table~\ref{tab:recorded-call-reuse} records invocations, not casewise
before/after migration pairs.
The later control study independently records successor \texttt{7804a781}
and calls the shared Hamiltonian and input-state pullbacks over each control
interval, as detailed in Appendix~\ref{app:protocol-bindings}. This connects
the package capability to the subsequent nominal and coupling-aware tasks.
The successor also changes a sparse JVP implementation; that separate path
is not used by the four-spin reverse-mode control study.

\paragraph{OpenPNM: completing transport parameter chains.}
The workflow evaluation reports 42 initially passing cases out of 80.
The extension is then revised to differentiate explicit linearized reaction
source arrays, retain separate directed-conductance components, and correct
flux cotangents and array-repetition transposes. The final 66 passes comprise
17 Stokes-flow, 17 Fickian-diffusion, eight heat-transport, 22
reaction--diffusion and two advection--diffusion cases. The latter two groups
account for the 24 additions. The remaining 14 capillary-invasion cases
request a hard-threshold invaded fraction and remain unsupported.

All 80 cases are run twice, with reference-value, derivative and duality
checks on the 66 passes. Small-step FD checks of the 22 reaction cases have
maximum reported relative discrepancy $9.15\times10^{-7}$; the much larger
original perturbations produce visible truncation error and are not used
as exact derivative references. Workflow code composes model construction,
transport and flux reduction using the package rules. The reported 42-to-66
result uses baseline \texttt{23587611} plus patch \texttt{ab59aaed\ldots}
under OpenPNM 3.6.4 against frozen 3.5.2 outputs. Retained revision
\texttt{d720f167} contains corresponding changes; byte identity with the
executed patch has not been established.

\paragraph{BioDynaMo and EvoEF2: preserving the computation around a rule.}
For BioDynaMo, the mechanics patch at \texttt{2938708} is evaluated in
descendant \texttt{65e3ab2}. It implements a Python sidecar for fixed-agent
contact mechanics; its value computation runs that kernel rather than
invoking the original C++ simulator. The report records eleven previously partial
cases becoming passes after matching the original sequential updates and
one-time Morton storage reorder. Local derivatives hold neighbor and
grid-box assignments fixed, with reference-value and JVP/VJP checks.
Its final 70 passes include 69 AD cases and one
forward-only case; ten discrete epidemic cases remain outside the AD result.
For EvoEF2 at \texttt{318c7b9}, original structure-preparation commands are
composed with derivatives of continuous scoring weights on the resulting
fixed structure. The report describes growth from 34 to 74 passing AD
cases: 22 mutant, ten repaired-structure and eight hydrogen-preparation
compositions are added to the evaluated task scope. The retained materials
establish these compositions, without identifying the research finding that
prompted the change or supplying paired earlier outputs.
Six branch-changing cases remain excluded; their
finite secants across discrete preparation choices are not local derivative
oracles.

\paragraph{QuantumOptics: expanding tasks with existing rules.}
The 49-task collection reports growth from 35 to 44 passes at
\texttt{1346bd2}. Nine added programs compose existing evolution,
steady-observation and spectral-projector rules: closed-system control,
correlation-spectrum fitting, Hamiltonian spectral inverse design,
jump-operator inference, open-system dynamic control, open-system Hamiltonian
inference, rate inference, semiclassical control and steady-state inverse
design. Inputs and numerical references remain fixed. Two complete runs
agree; the maximum reported independent-FD discrepancy is
$1.75\times10^{-7}$. Five tasks remain unsupported where the required
correlation, stochastic-path or model-generation chain is absent.
These tests evaluate nine new compositions of existing rules through their
numerical outputs and derivatives. Optimizer completion is a separate test.
Appendix~\ref{app:task-correspondence} describes the separately measured
144-control gradient as a concrete example of shared-rule composition.

\section{Quantum control and phase estimation}
\label{app:quantum-control}

The control study first constructs nominal probes and evaluates their
sensitivity to coupling errors. It then compares nominal and revised
objectives with matched evaluation budgets. The final subsection defines
the measurement task that uses saved states from this comparison.

\paragraph{Physical problem and active controls.}
With $\hbar=J=1$, we use the full Hilbert space of four spin-$1/2$ sites,
open boundaries and Pauli matrices:
\[
    H_k=-\sum_{i=0}^{2}Z_iZ_{i+1}
    +\sum_{i=0}^{3}\bigl(u_{k,i,x}X_i+u_{k,i,y}Y_i\bigr).
\]
Time is measured in $J^{-1}$. There are $M=12$ equal-duration intervals
of length $T/12$, each with eight real amplitudes
bounded by $|u_{k,i,a}|\leq2$. The initial state is
$|\uparrow\uparrow\uparrow\uparrow\rangle$ and the target is
$|\mathrm{GHZ}_+\rangle=(|\uparrow\uparrow\uparrow\uparrow\rangle+
    |\downarrow\downarrow\downarrow\downarrow\rangle)/\sqrt{2}$.
The 96-dimensional optimization objective is
\[
    \mathcal{J}(u)=1-|\langle\mathrm{GHZ}_+|\psi(T)\rangle|^2
    +10^{-3}\frac{1}{8M}\sum_{k,i,a}u_{k,i,a}^2 .
\]
The penalty is mean squared control amplitude, not integrated pulse energy.
Durations are $T\in\{2,4,6\}$. For each duration, seeds 0--4 define uniform
initial amplitudes in $[-0.3,0.3]$; the same seed uses the same initial pulse
across durations. L-BFGS-B uses a 200-iteration cap,
$\texttt{gtol}=10^{-6}$ and $\texttt{ftol}=10^{-12}$.

\paragraph{Composition and numerical checks.}
The frozen extension commit is \texttt{7804a781}, with QuSpin 1.0.1 and
the package's declared Python ChainRules 0.1.0 dependency. Native
\texttt{hamiltonian.evolve} supplies each interval's forward state.
The registered pullback returns both the Hamiltonian cotangent and the
initial-state cotangent, which is propagated to the preceding interval.
For $c=\langle\mathrm{GHZ}_+|\psi(T)\rangle$, the terminal seed is
$-2c|\mathrm{GHZ}_+\rangle$ under the real pairing
$\operatorname{Re}\langle \bar\psi,\delta\psi\rangle$.
Contractions with the physical X/Y operators give the real control
gradient; the task adds only the explicit quadratic penalty.

Before optimization, two- and four-spin three-interval probes check every
one of 12 and 24 control coordinates against native-value central FD.
The subsequent 96-control runs check nonzero directions before optimization
and at the terminal pulse with steps $10^{-4},10^{-5},10^{-6}$.
Terminal checks use the bare infidelity gradient to avoid a vacuous
near-zero total-gradient comparison at a regularized stationary point.
Native propagation is compared with tightened solver tolerances and dense
matrix exponentials. A separate checker, importing neither QuSpin nor the
task runner, reconstructs tensor-product Pauli Hamiltonians from saved
controls. It reproduces all 15 terminal states to
$8.76\times10^{-12}$ and repeats the recorded directional checks, adding an
independent random direction at each terminal pulse.

\paragraph{Optimization outcomes.}
The fidelity range is $0.9999947348$--$0.9999999653$.
All optimizers return the iteration-limit status, with projected-gradient
infinity norms from $4.63\times10^{-6}$ to $1.54\times10^{-4}$.
The complete controls and termination records are preserved. The
96-control derivative checks sample directions rather than every coordinate;
this study does not measure an AD/FD optimization time ratio.

\paragraph{Frozen-pulse evaluation.}
After optimization, every pulse is evaluated at
$H_k(\delta)=(1+\delta)H_0+\sum_{i,a}u_{k,i,a}\sigma_i^a$
with $H_0=-\sum_{i=0}^{2}Z_iZ_{i+1}$, for
$\delta=-0.10,-0.08,\ldots,0.10$, without changing its controls.
The zero-error point is the nominal reference; the other ten values were
not used for optimization. The error multiplies only the Ising drift; it
does not perturb control amplitudes or interval durations.
All $15\times11=165$ pulse trajectories and
$3\times11=33$ zero-control references are retained.
Across their 2,376 interval evolutions, native and dense states agree to
$1.29\times10^{-12}$. The zero-control fidelity is $1/2$ for every duration
and error, as expected for the initial eigenstate of the Ising drift.

The representative is selected by the smallest nominal regularized
objective among the five $T=4$ runs before error evaluation: seed 3.
Its half-chain purity changes from 1 to 0.4999998199 and
$S_2=-\log_2\operatorname{tr}(\rho_A^2)$ approaches one bit.
The tensor mapping is checked using the original basis labels and each
site's Z operator; states are not renormalized to pass the checks.
High GHZ overlap, rather than entropy alone, identifies the target state.

\begin{figure}[!htbp]
    \centering
    \includegraphics[width=\linewidth]{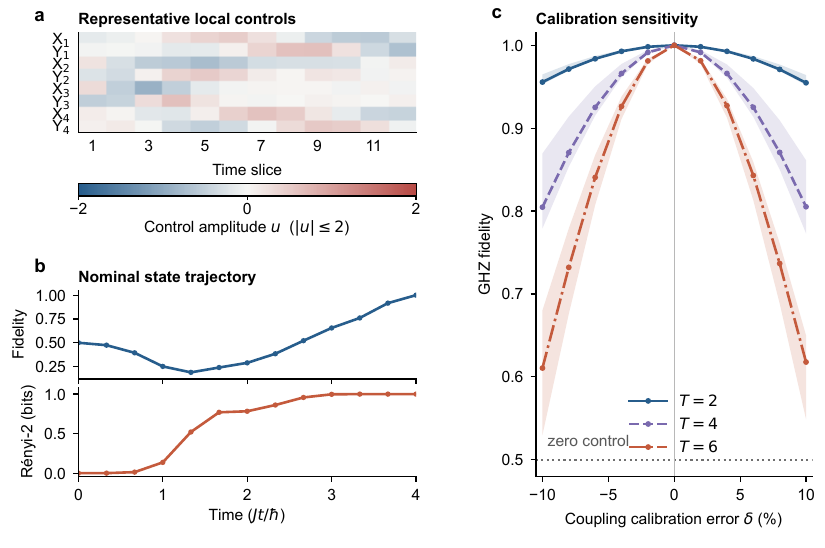}
    \caption{Complete control and evaluation record for the four-spin study.
        (a) The representative's 96 physical control amplitudes.
        (b) GHZ fidelity and half-chain R\'enyi-2 entropy along its nominal evolution.
        (c) Fixed-pulse response to the specified coupling-error grid, with the
        median and observed range across all five initializations at each duration.
        All evaluated coupling offsets are shown.}
    \label{fig:control-detail}
\end{figure}

The records include all 15 optimizations and 198 subsequent propagation
trajectories. Independent Pauli-matrix reconstruction requires NumPy/SciPy;
rerunning the optimizations uses the recorded QuSpin/ChainRules environment
and a dense reverse rule. Additional
time-discretization and six-spin checks, and a single-direction sparse-JVP
probe for single-excitation transport \citep{coopmans2022disorder}, are
documented with the code and data. These probes evaluate time resolution,
system size and a single-direction sparse derivative separately.

\subsection{Revising the control objective after calibration-error evaluation}
\label{app:control-robust}

The earlier fixed-pulse evaluation motivates a comparison at $T=4$ with
the same four-spin model, 12 intervals, 96 controls and package derivatives.
Writing $\ell_\delta(u)=1-|\langle\mathrm{GHZ}_+|\psi_\delta(T)\rangle|^2$,
the two objectives are
\[
    J_N(u)=\ell_0(u)+10^{-3}\operatorname{mean}(u^2),\qquad
    J_R(u)=\frac13\!\sum_{\delta\in\{-0.1,0,0.1\}}\!\ell_\delta(u)
    +10^{-3}\operatorname{mean}(u^2).
\]
The revised objective averages three scenarios; it is not minimax.
Their error values were already observed in the earlier study. Before new
optimization, we fix ten additional confirmation points
$\{-0.09,-0.07,\ldots,0.09\}$, which do not overlap the old grid.
No design is revised after these points are evaluated.

The seeds 20260922--20260924 generate three new uniform initial pulses in
$[-0.3,0.3]^{96}$ using NumPy's default generator, paired between objectives.
They are not warm starts from previous optimized designs. L-BFGS-B uses
the same bounds and penalty, $\texttt{gtol}=10^{-6}$,
$\texttt{ftol}=10^{-12}$, $\texttt{maxcor}=10$ and $\texttt{maxls}=20$.
Each arm receives 660 single-scenario value-and-gradient evaluations per
seed: the nominal objective consumes one per request, and the three-scenario
objective consumes three. All six runs exhaust this budget before their
1,000-iteration cap. They return the last accepted iterate, not a point
selected using confirmation results. Actual accepted iterations are
618/617/621 for nominal control and 213/204/199 for the three-scenario arm;
equal budgets therefore do not mean equal iterations.

\begin{figure}[!htbp]
    \centering
    \includegraphics[width=0.60\linewidth]{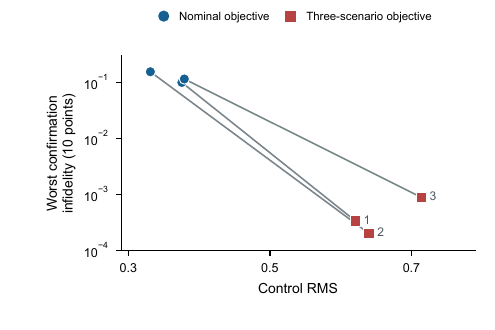}
    \caption{Control-strength cost of reduced preparation error.
        Each line joins nominal and three-scenario designs from the same
        initialization. Across ten fixed confirmation points, the three-scenario
        designs have lower worst infidelity and higher control RMS in all three
        pairs. Lines identify paired initializations.
        All runs receive the same 660 single-scenario evaluation budget;
        Figure~\ref{fig:evolution}b gives their individual error evaluations.}
    \label{fig:control-robust}
\end{figure}

Across the three pairs, minimum confirmation fidelity is 0.898473,
0.842458 and 0.882875 for nominal control, and 0.999657, 0.999797 and
0.999100 for the three-scenario designs. RMS control amplitudes increase
by 0.24494, 0.30902 and 0.33491, while nominal fidelity decreases by
$3.49\times10^{-4}$, $2.06\times10^{-4}$ and $9.20\times10^{-4}$.
Figure~\ref{fig:control-robust} shows every seed rather than a selected
best run. These are finite-grid trade-offs, not a continuous-error guarantee.

All 78 final trajectories---six designs at three training and ten confirmation
points---are checked against independent matrix exponentials. An additional
review reconstructs the Pauli Hamiltonians without importing QuSpin or the
research program: its maximum fidelity difference is $2.29\times10^{-13}$,
and six new directional FD checks agree with saved gradient contractions
to $3.61\times10^{-11}$. Production gradients still use the original
package pullbacks; the independent exponentials are only validation references.
At budget exhaustion, projected gradient norms are
$7.12\times10^{-6}$--$5.06\times10^{-4}$. The existing package and Python
ChainRules 0.1.0 are reused with QuSpin 1.0.1, Python 3.12.13,
NumPy 2.5.2 and SciPy 1.18.1. No AD/FD timing comparison is made.

\subsection{Fixed-readout finite-shot phase estimation}
\label{app:quantum-metrology}

The sensing task estimates a dimensionless collective phase from repeated
measurements of a prepared four-spin state. First, the frozen control pulse
produces $|\psi_u(\delta)\rangle$ under the preparation Hamiltonian above.
Then an unknown phase $\phi$ is encoded by $G=\tfrac12\sum_j Z_j$.
A known bias $\beta=\pi/8$ and a tensor-Hadamard rotation $R$ define the fixed
readout, with all 16 bit strings retained:
\[
    p_z(\phi\mid u,\delta)=
    |\langle z|R\exp[-i(\phi+\beta)G]|\psi_u(\delta)\rangle|^2.
\]
The initial readout evaluation covers all six controls, the 13 existing
coupling conditions and 41 phases in $[-\pi/16,\pi/16]$. The coupling error
is fixed and known to the estimator; unknown drift and measurement noise
are different inference problems.

This evaluation distinguishes information in the state,
$F_Q=4\operatorname{Var}(G)$, from information accessible through the chosen
readout, $F_C=\sum_z(\partial_\phi p_z)^2/p_z$.
The smallest nominal $F_C/F_Q$ is 0.3114, at seed 20260923,
$\delta=0.1$, $\phi=-\pi/16$; the robust controls remain above 0.9983
on this evaluated grid. Probabilities and their phase derivatives are
retained and checked independently. These quantities motivate testing
finite measurement budgets, without identifying a Fisher bound with an
estimator's observed error.

The finite-sample illustration selects the nominal pulse from that seed and
one robust pulse from seed 20260924, both at $\delta=0.1$.
The nominal pulse was selected using the preceding information analysis;
the two controls are not paired initializations. All sampling conditions
were fixed before generating observations. The phases tested here are
$0$ and $\pi/32$, rather than the earlier minimum-information phase.
An ideal unentangled state $|+\rangle^{\otimes4}$ and an ideal GHZ state
provide reference probes under the same encoding and readout. They are
specified directly and have no matched preparation-cost measurement.

Each simulated experiment draws a 16-outcome multinomial count vector from
$\nu\in\{100,1000,10000\}$ measurements and estimates $\phi$ by maximizing
its likelihood over $[-\pi/16,\pi/16]$. Each of the 24 state/phase/budget
conditions has 500 independent repetitions. An 8,193-point likelihood grid
is checked against 32,769 points on 20 repetitions per condition; their
maximum estimate difference is $2.40\times10^{-5}$ radians. There are no
estimator failures; the ideal unentangled reference reaches the interval
boundary in 1.8\% of repetitions at 100 measurements and $\phi=\pi/32$.

Figure~\ref{fig:evolution}c shows $\sqrt{\operatorname{mean}
        [(\hat\phi-\phi)^2]}$ at 10,000 measurements, with approximate Monte Carlo
standard errors obtained from the standard error of the mean squared error
(MSE) by the delta
method. The two prepared states have errors of roughly 2.5--2.9 milliradians,
near the ideal GHZ reference and below the 4.8--5.1 milliradians of the
unentangled reference. The nominal-to-robust RMSE reductions at the two
phases are 13.4\% and 2.5\%; the latter difference is smaller than the
combined Monte Carlo standard error. Consequently this selected comparison
illustrates sensing utility, rather than establishing a general gain from
the robust objective. The full records retain bias, MSE, uncertainty and
the separate $1/(\nu F_C)$ reference for all three measurement budgets.

\section{FiPy cloak design and material-error analysis}
\label{app:cloak}

The multiload study supports the design revision and final evaluation in
Section~\ref{sec:evolution}. A separate material-error study tests a further
objective revision under conductivity tolerances, retaining the same
diffusion derivative.

\subsection{Passive multiload design and task revision}

The application in Figure~\ref{fig:research-outcomes} uses eight equal-width
layers between radii 0.16 and 0.30 on the unit square, with core conductivity
0.05 and background conductivity one. It optimizes eight log-conductivities,
subject to $0.05\leq k_i\leq8$ and area-weighted mean at most 2.5.
The steady field solves $-\nabla\cdot(k\nabla T)=0$, with prescribed
temperatures on all four outer sides. Let $T_\ell^{\rm hom}$ and
$T_\ell^{\rm bare}$ denote the homogeneous-background and bare-core
solutions for loading $\ell$ on the same grid. The objective is
\[
    J(\theta)=\frac{1}{|\mathcal L|}\sum_{\ell\in\mathcal L}
    \frac{\operatorname{mean}_{c\in\Omega_{\rm obs}}
    [T_\ell(c;\theta)-T_\ell^{\rm hom}(c)]^2}
    {\operatorname{mean}_{c\in\Omega_{\rm obs}}
    [T_\ell^{\rm bare}(c)-T_\ell^{\rm hom}(c)]^2},
    \qquad \Omega_{\rm obs}=\{0.34<r<0.46\}.
\]
Each load is normalized before averaging; prescribed boundary temperatures
are not scored. Conductivities are constant in the fixed layers,
$k_i=e^{\theta_i}$. Interior face coefficients use
$k_f=2k_Lk_R/(k_L+k_R)$, while boundary faces use their adjacent cell value.
The material budget uses training-grid shell-cell area fractions.
The original FiPy solve and the extension's implicit derivative are composed
through Autograd; no new diffusion derivative is implemented by the task.

With $z=(x-\tfrac12)+i(y-\tfrac12)$, the horizontal, quadratic and cubic
Dirichlet profiles are $\tfrac12+\operatorname{Re}z$,
$\tfrac12+2\operatorname{Re}z^2$ and
$\tfrac12+2\operatorname{Re}z^3$. The analytic uniform-shell reference
uses the positive root of
\[
    (1-f)k_s^2+(1+f)(k_c-1)k_s-(1-f)k_c=0,
    \qquad f=(0.16/0.30)^2,\quad k_c=0.05,
\]
giving $k_s=1.73411269$. The optimized uniform and equal-thickness two-layer
references use the same $192^2$ training grid, three-load objective,
conductivity bounds and material budget as the eight-layer design, with
FiPy AD gradients and SLSQP tolerance $\texttt{ftol}=10^{-11}$.
Each reference starts uniformly at $k_s$. The eight-layer design additionally
uses an alternating $k_s e^{-0.8}/k_s e^{0.8}$ start and selects the feasible
terminal design with the lower training objective. Thus the physical
objective and constraints match, but the number of optimization starts does not.
The optimized uniform conductivity is 1.05688113; the two-layer values are
2.49493962 and 0.84793788 from inner to outer, meeting at $r=0.23$.

The first design trains on horizontal and quadratic boundary profiles on a
$96^2$ grid. Its cubic-load disturbance on the common $192^2$ evaluation
grid is 2.956 times that of the bare inclusion. The revision adds the cubic
profile to training, uses a $192^2$ training grid and selects a feasible
SLSQP design from two starts. At $192^2$, the revised cubic ratio is 0.003713.
Because the loading has now been used in development, this is a repaired
training condition, not unseen-load generalization; changing the training
grid and objective together also prevents attributing the improvement to one
change alone. The selected run reports successful termination at 160 iterations.

On the $576^2$ evaluation grid, reductions in exterior RMS disturbance are
97.05\%, 99.20\% and 95.54\% for horizontal, quadratic and cubic loads;
a new localized heating load improves by 96.89\%. The quartic challenge
instead has 2.777 times the bare-inclusion disturbance. The analytic uniform
coating outperforms the eight-layer design for horizontal loading, while the
optimized two-layer baseline is better for the cubic loading. Thus the design
is a constrained multiload compromise, not a uniformly superior cloak.
All controls and load results are retained with the figure source data.
The displayed temperature fields and quantitative load comparisons both
use $576^2$ cells, with the optimized conductivities held fixed.
The three displayed forward fields were checked against an independent
finite-volume solve, with maximum temperature difference $3.29\times10^{-12}$.

For each loading define the RMS reduction
$R_{\ell,n}=1-\operatorname{RMSE}_{\ell,n}^{\rm cloak}/
    \operatorname{RMSE}_{\ell,n}^{\rm bare}$.
The declared grid-stability test requires
$\max_\ell |R_{\ell,576}-R_{\ell,384}|<0.03$ across all eight training and
evaluation loads. Besides the five loads in Figure~\ref{fig:research-outcomes},
this set includes vertical heating, a rotated quadratic profile and an
earlier localized-heating profile. These additional profiles were already
used in the initial design's evaluation; they are distinct from the new
profiles in the material-error study below. The thermal-cloak design
protocol and grid-refinement data list the complete set.
The test fails because the quartic-load reduction changes by
3.35 percentage points. A later fixed-design $768^2$ check gives a maximum
change of 1.68 points relative to $576^2$; it supplements rather than replaces
the failed test.

\subsection{Material-error follow-up}

\paragraph{Problem and comparison.}
Conductivity tolerances can change the behavior of an optimized coating.
This follow-up reoptimizes both designs; they are distinct from the fixed
design evaluated in Figure~\ref{fig:research-outcomes}.
The material-error study tests a design objective that includes these errors,
using the same geometry, steady diffusion equation, observation annulus and
three-load normalization as above. Its training-grid improvement does not
persist under finer evaluation, illustrating why the revised objective also
needs a discretization check.
For each material-error scenario $s$, $J_s$ is the mean of these normalized
\emph{squared} errors over three training loads: linear, quadratic and cubic
harmonics. The nominal arm minimizes $J_0$; the finite-scenario arm minimizes
$\max_sJ_s$, where the five scenarios are zero error, common $\pm5\%$ error,
and alternating-layer $\pm5\%$ error. This is a worst material scenario
of a three-load mean, not the worst individual load or the whole error box.

Coordinates and conductivities are normalized to the square side length and
background conductivity; boundary temperatures use the profiles below.
Both arms constrain $0.05/0.95\leq k_i\leq8/1.05$ and impose
area-weighted mean conductivity at most $2.5/1.05$, using each of the exact
annular areas and the $192^2$ and $384^2$ pixel areas. These constraints keep
all componentwise $\pm5\%$ material perturbations feasible.
The shared starts are a uniform conductivity of 1.73411269 and an
alternating $e^{-0.8}/e^{0.8}$ multiple of that value, rescaled to the area
budget. SLSQP uses $\texttt{ftol}=10^{-11}$ and a cumulative 180-iteration
cap, with a limit of 3,000 forward plus 3,000 reverse solves per trajectory.
The two nominal runs converge; both finite-scenario runs exhaust their
solve budget. Uniform-start runs resume from saved parameters with cumulative
budgets preserved. The original design-export routine incorrectly used an
auxiliary upper-bound variable for the training loss to screen material-feasible cached points.
A post-run correction selected each arm's lowest-training-loss feasible point
from its unchanged cache and computed a consistent upper-bound value from that
loss, preserving the original records.
Only values and gradients were rechecked; no further optimization or selection
on evaluation results was performed. The comparison therefore uses feasible
designs within the recorded budgets, not converged optima for both objectives.
The original caches and selection record preserve this correction.
The extension is FiPy AD 0.2.1, commit \texttt{f7314e0d}, with the original
FiPy forward solver. The task defines perturbations and loss aggregation,
without modifying package code or implementing a replacement derivative.

The three training profiles are those defined above. The two previously
specified challenge loads
are $1/2+4\operatorname{Re}z^4$ and
$1/2+(1/2)\exp[-((x-1)/.10)^2-((y-.72)/.18)^2]$.
The three new profiles use $1/2+8\operatorname{Re}z^5$ and localized
hotspots
$1/2+(1/2)\exp[-((x-.73)/.11)^2-(y/.09)^2]$ and
$1/2+(1/2)\exp[-(x/.08)^2-((y-.24)/.13)^2]$.
These functions prescribe Dirichlet data on all four sides.

\paragraph{Independent evaluation.}
With both designs fixed, evaluation uses 21 material settings and five known
loads at $192^2$, and the five training material settings with those loads at
$384^2$. Both grids also use 12 predeclared material-error vectors and the
three new loads. An independent finite-volume assembly checks the fine-grid
results: 30 native-versus-independent comparisons have maximum temperature
difference $6.06\times10^{-13}$. Every checked field satisfies the prescribed
heat-balance tolerance. All material/load results are retained.

\begin{table}[htbp]
    \centering
    \small
    \caption{Thermal-cloak objective trade-off. $J_0$ is the nominal mean
        normalized squared error; $\max_s J_s$ uses the five training material
        settings. Both designs remain frozen across grids. Lower is better.}
    \label{tab:cloak-objectives}
    \begin{tabular}{@{}llrr@{}}
        \toprule
        Grid    & Design objective      & $J_0$               & $\max_sJ_s$ \\
        \midrule
        $192^2$ & Nominal               & $6.42\times10^{-6}$ & 0.48828     \\
        $192^2$ & Five-scenario minimax & 0.001209            & 0.46369     \\
        $384^2$ & Nominal               & 0.001578            & 0.51741     \\
        $384^2$ & Five-scenario minimax & 0.004002            & 0.53757     \\
        \bottomrule
    \end{tabular}
\end{table}

\paragraph{Material robustness across grids.}
The training-grid worst-scenario loss decreases by about 5\%, but its
ordering reverses on the finer grid (Table~\ref{tab:cloak-objectives}).
Under common material errors, the cubic-load disturbance still exceeds
that of the bare disk for both designs on both grids. In the new tests,
both localized hotspots have small relative disturbances, whereas the
fifth harmonic remains worse than the bare disk. At $384^2$, the largest
hotspot RMSE/bare-disk ratios are 0.0535 and 0.0624 for the nominal and
minimax designs; fifth-harmonic maxima are 30.8 and 38.3.
The latter ratios have small bare-disk denominators: their corresponding
absolute RMSE maxima are $2.05\times10^{-4}$ and $2.55\times10^{-4}$.
Absolute values and all individual conditions are retained, rather than
interpreting the ratios as large absolute temperature errors.
For these frozen, budget-limited designs, the data do not separate incomplete optimization from the
effects of the finite training conditions and discretization.

\section{Inverse-problem definitions and results}
\label{app:checks}
\label{app:inverse}
\label{app:numerical-validation}
\label{app:full-research-plots}

\subsection{Thermal conductivity inversion}

The thermal experiment uses FiPy AD 0.2.1 with Autograd 1.8.0 and the original
FiPy SciPy solver backend. It runs on Apple M4 hardware, macOS 26.6.2 arm64,
Python 3.12.14, NumPy 2.5.3 and SciPy 1.18.1. Numerical thread limits are
one; processes run serially without CPU affinity or background-load isolation.
Three complete inversions per method provide the timing comparison.

The unit square has $32\times32$ cell-centred finite volumes and $8\times4$
conductivity zones. At cell centre $(x,y)$, let
$z=8\lfloor4y\rfloor+\lfloor8x\rfloor$ and $k=\exp\theta_z$; face coefficients
use arithmetic interpolation. At zone centres, the synthetic log-conductivity is
\begin{align*}
    \theta^\star(x,y)={} & 0.70\exp\!\left[-\frac{(x-.28)^2+(y-.33)^2}{.035}\right]
    -0.65\exp\!\left[-\frac{(x-.72)^2+(y-.68)^2}{.040}\right]                       \\
                         & +0.15\sin(2\pi x)\cos(\pi y).
\end{align*}
Each loading solves $-\nabla\cdot(k\nabla T)=q$. Training loading 1 sets
$T=1$ on the left and $T=0$ on the right, zero flux on the other sides,
and $q=2$. Loading 2 sets $T=0$ on the bottom and $T=1$ on the top, zero
flux on the other sides, and
$q=1+4\exp[-((x-.65)^2+(y-.30)^2)/.035]$.
The held-out loading sets $T=0$ on all sides and
$q=15\exp[-((x-.40)^2+(y-.65)^2)/.025]$.
Original FiPy produces noiseless temperatures at every cell centre for
each loading. The loss is
$100\operatorname{mean}_{m,c}(T_{m,c}(\theta)-T^\star_{m,c})^2$
over the two training loadings and 1,024 cells; it has no regularizer.
Both methods start at $\theta=0$ with bounds $[-1,1]$.
The forward solver is FiPy's SciPy LinearLUSolver (tolerance $10^{-12}$,
iteration limit 1,000). L-BFGS-B uses \texttt{maxiter=250},
\texttt{maxls=30}, \texttt{maxcor=20}, \texttt{ftol=1e-14}, and
\texttt{gtol=1e-8}; centred FD uses step $10^{-5}$.
After optimization, original FiPy recomputes all three fields. The checks
require optimizer success, a held-out RMSE below 1\% of its initial value,
and relative conductivity error below 1\%. This is a same-grid synthetic
inversion with full-field observations, not a noisy or sparse-sensor test.
The recorded AD optimization takes 134 iterations and 145 objective-gradient
evaluations; FD takes 124 iterations and 132 evaluations.
The archive records zero maximum forward discrepancy and relative L2
gradient discrepancy $4.80\times10^{-9}$ against FD.
Relative conductivity errors are $3.07\times10^{-6}$ for AD and
$3.46\times10^{-6}$ for FD. AD's held-out temperature RMSE is
$1.72\times10^{-7}$, or $3.18\times10^{-5}$ of the initialization error.
Each FD value-and-gradient evaluation uses $2P+1=65$ objective calls and two
forward solves per call: 132 evaluations therefore require 17,160 solves,
versus AD's 290 forward solves plus the adjoint work included in its wall time.
Including setup and post-fit validation changes the optimization-only
39.5-fold time ratio to a 28.6-fold complete-process ratio.

\begin{figure}[!htbp]
    \centering
    \includegraphics[width=\linewidth]{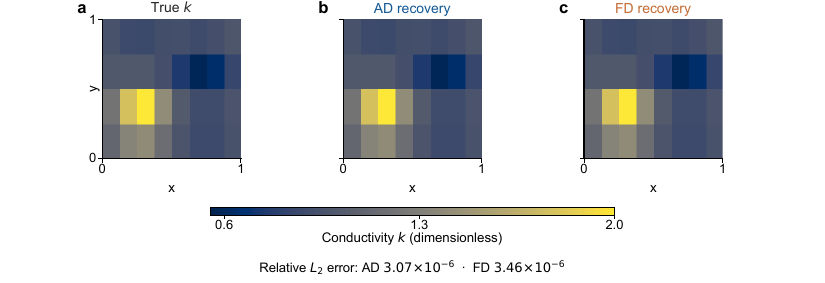}
    \caption{Recovery of the 32-region FiPy conductivity field. True conductivity
        and the AD and FD recovered fields use the same dimensionless color scale
        and original $8\times4$ parameter zones. Relative $L_2$ errors are
        $3.07\times10^{-6}$ and $3.46\times10^{-6}$ for AD and FD. These are the
        synthetic inverse-problem results, separate from thermal-cloak design.}
    \label{fig:fipy-recovery}
\end{figure}

\subsection{Local-density-of-states inversion}

The Kwant experiment uses AD version 0.2.1.dev1, Kwant 1.5.0 with its SciPy
backend, Python 3.12.14, NumPy 1.26.4 and SciPy 1.13.1 on Apple M4 hardware
with 16 GiB memory. Numerical thread limits are one and three processes per
method run serially, without CPU affinity or background-load isolation.
The fixed source and wheel identities match the execution records. This
inversion is separate from the eight-gate filter design in
Appendix~\ref{app:transport-design}.

The model is a one-orbital, 36-site chain with hopping $-0.91$.
For $x=0,\ldots,35$, its onsite potential is $0.07\cos(0.37x)$, plus
$\phi_{x-2}$ on the 32 central sites $2\leq x<34$.
Each end attaches to a semi-infinite chain with onsite zero and hopping $-1$.
The synthetic truth is
$\phi_i^\star=.035\sin(.43i+.2)+.018\cos(.77i-.1)$, $i=0,\ldots,31$.
Original Kwant generates noiseless LDOS at every site at training energies
$(-.9,-.35,.15,.65)$ and held-out energies $(-.55,.37)$.
With $D^\star$ the $4\times36$ training array and
$s=\sqrt{\operatorname{mean}[(D^\star)^2]}$, the loss is
\[
    L(\phi)=\operatorname{mean}\left[\left(\frac{D(\phi)-D^\star}{s}\right)^2\right]
    +10^{-6}\operatorname{mean}(\phi^2).
\]
Both methods start at zero with bounds $[-.15,.15]$. L-BFGS-B uses
\texttt{maxiter=180}, \texttt{maxls=30}, \texttt{ftol=1e-13}, and
\texttt{gtol=1e-7}; centred FD uses step $10^{-5}$.
The reference calls Kwant's public sparse LDOS solver with its default
solver settings. AD sums one VJP per training energy and adds the ridge
gradient. After optimization, original Kwant checks the fitted and held-out
LDOS; both residual-norm ratios must be below 0.1 relative to initialization.
An independent FD recomputation also checks
$\|\phi-\operatorname{clip}(\phi-\nabla L,-.15,.15)\|_\infty\leq2\times10^{-5}$.
The post-fit projected FD gradient infinity norm is
$7.93\times10^{-8}$; this assesses stationarity, not derivative accuracy.
These parameters are in the benchmark's numerical energy convention;
no conversion to a specified material's physical units is claimed.
Both methods reach final losses of approximately $6.72\times10^{-10}$.
The AD parameter-recovery relative error is $1.33\times10^{-4}$;
prediction RMSE over the 72 withheld site/energy values is
$3.87\times10^{-7}$. The separate four-size Kwant gradient comparison has
maximum absolute component discrepancies between $1.48\times10^{-12}$ and
$5.90\times10^{-12}$; these derivative checks are distinct from inverse-solution
stationarity and parameter recovery.

\subsection{Prediction, recovery and optimization cost}

FiPy's time curves use accepted iterates and their recorded wall times,
normalizing each objective by its initial value. The untimed iteration-zero
anchor is omitted from the logarithmic axis, while complete-run summaries
include the whole optimization interval, excluding setup and post-fit validation.
Kwant supplies total optimizer
times but no per-iteration time record, so no time trajectory is constructed.
Median optimization times for AD and FD are 1.70 and 67.3\,s in FiPy,
and 2.61 and 76.9\,s in Kwant, giving FD/AD time ratios of 39.5 and 29.5.
The three FiPy runs span 1.69--1.87 s for AD and 64.6--72.9 s for FD;
Kwant spans 2.50--2.67 s and 75.0--86.2 s. These are observed ranges,
not confidence intervals or paired speedup intervals.

For value-and-gradient calls with $N_{\mathrm{FD}}=2P+1$ and
$\widehat c$ as defined in Equation~\ref{eq:cost},
the optimization-time ratio can be decomposed as
\[
    \frac{T_{\mathrm{FD}}}{T_{\mathrm{AD}}}
    \approx\frac{n_{\mathrm{FD}}}{n_{\mathrm{AD}}}
    \frac{2P+1}{\widehat c}.
\]
FiPy uses 132 FD versus 145 AD evaluations, with two thermal loads in each
objective; its aggregate per-evaluation proxy is about 1.50.
Kwant uses 100 evaluations per method and four training energies, giving
a proxy of 2.20. These workload definitions differ from the single-energy
Kwant gradient-scaling series.

\begin{figure}[!htbp]
    \centering
    \includegraphics[width=\linewidth]{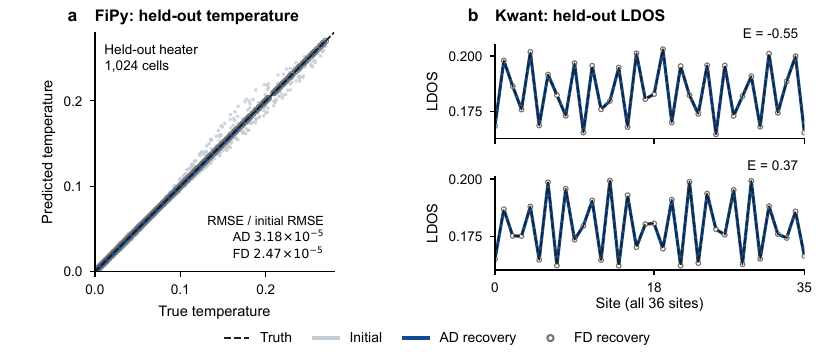}
    \caption{Predictions at conditions excluded from fitting.
        (a) All 1,024 temperatures under FiPy's withheld heater loading are compared
        with their synthetic targets. (b) Kwant LDOS profiles at energies $-0.55$ and
        $0.37$ include all 36 sites, giving 72 withheld outputs. Both methods use the
        same model and discretization as the synthetic data. These checks test the
        fitted parameters at additional conditions, not robustness to model error.}
    \label{fig:heldout-validation}
\end{figure}

\begin{figure}[!htbp]
    \centering
    \includegraphics[width=\linewidth]{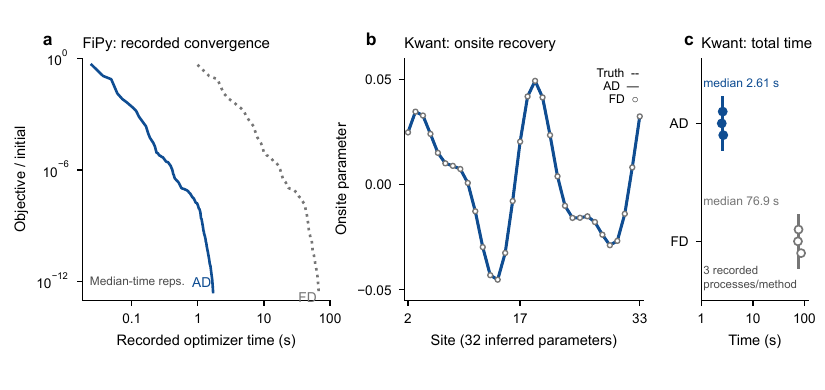}
    \caption{Additional records for the two synthetic inverse problems.
        (a) FiPy objective relative to its initial value versus recorded optimizer
        time, using each method's median-time process. The untimed initial value is
        omitted from the logarithmic time axis.
        (b) True and recovered onsite parameters at the 32 inferred sites of the
        Kwant LDOS problem.
        (c) All three Kwant optimizer times per method; vertical marks denote
        medians. Only total optimizer times, not iteration-level times, were recorded
        for Kwant. Held-out predictions appear in Figure~\ref{fig:heldout-validation}.}
    \label{fig:applications}
\end{figure}

The displayed fields use the representative processes in
Figure~\ref{fig:applications}. FiPy's held-out temperature RMSE is
$1.72\times10^{-7}$ for AD and $1.34\times10^{-7}$ for FD, or
$3.18\times10^{-5}$ and $2.47\times10^{-5}$ of the initial error.
Kwant's withheld-output RMSE is $3.87\times10^{-7}$ and
$3.89\times10^{-7}$, respectively. The comparison therefore pairs
computational cost with similar recovery and prediction quality on the
specified synthetic problems.

\end{document}